\documentclass{article}
\usepackage{preprint,times}

\usepackage{hyperref}
\usepackage{url}
\usepackage{amsmath}
\usepackage{amssymb}
\usepackage{adjustbox}
\usepackage{array}
\usepackage{booktabs}
\usepackage{float}
\usepackage{graphicx}
\usepackage{longtable}
\usepackage[section]{placeins}
\usepackage[most]{tcolorbox}
\usepackage{hyperref}
\usepackage{url}
\usepackage{enumitem}

\newtcolorbox{benchmarkbox}[1][]{
  enhanced,
  colback=black!1,
  colframe=black!45,
  boxrule=0.45pt,
  arc=1mm,
  borderline west={1.2pt}{0pt}{black!55},
  left=6pt,
  right=6pt,
  top=3pt,
  bottom=3pt,
  before skip=2pt,
  after skip=1pt,
  #1
}

\title{Echoes of Deeds: Moral History Can Shape and Steer LLM Behavioral Choices}

\author{Lucio La Cava\\
DIMES Dept., University of Calabria, Italy\\
\texttt{lucio.lacava@dimes.unical.it}
\And
Andrea Tagarelli \\
DIMES Dept., University of Calabria, Italy\\
\texttt{tagarelli@dimes.unical.it}
}

\iclrfinalcopy

\begin{document}

\maketitle

\begin{abstract}
Evaluations of Large Language Models (LLMs) morality typically consider decisions in isolation, thus overlooking whether an individual's unrelated prior conduct  influences the model's subsequent choices. 
This leaves open the question of whether, and to what extent, moral history shapes LLM decisional behaviors. 
Prior work on human moral decision-making shows that past behavior can influence subsequent moral choices. Building on this observation, we investigate whether analogous effects emerge in LLMs in two complementary ways:  
at the behavioral level, through the model's observable responses, and at the representation level, through its latent internal representations.
We introduce MoralLedger, a framework for studying how an actor's moral history shapes actions for LLMs' behaviors under a fixed decision context. 
At the behavioral level, we find that prior moral histories systematically alter subsequent choices as a function of their valence and intensity. At the internal representation level, these histories induce a linearly recoverable direction in the residual stream that generalizes to held-out examples. Intervening along this direction on neutral-history prompts produces two-sided intensity-dependent changes in subsequent choices, with effects that are stronger than those induced by prompting alone or by favorable-nonmoral direction. 
To our knowledge, this is the first demonstration that a latent representation of an actor's prior moral conduct can provide signed inference-time control over a moral decision. 
Our MoralLedger extends moral evaluation beyond static dilemmas, establishing moral history as both a source of behavioral sensitivity and a causal target for auditing and controlling moral behavior in LLMs.
\end{abstract}

\section{Introduction}
\vspace{-1.5mm}
In humans, moral choices are not always evaluated in isolation, as prior conduct has been shown to systematically influence subsequent behavior: past actions might promote behavioral consistency, license subsequent self-interest, or motivate compensatory behaviors after wrongdoing~\citep{mullen2016consistency,ferguson2024influences}. These effects highlight that human moral behavior can depend not only on the decision currently being faced, but also on the moral history that precedes it.
As a result of their training, LLMs may internalize narrative structures that reflect such moral patterns~\citep{emelin-etal-2021-moral,SchramowskiTARK22}. Indeed, their behavior at inference is sensitive to contextual descriptions, narrative framing, and persona information~\citep{choi2024picle}. 

The above raises a key question: 
\textit{Can an actor's prior moral conduct systematically influence an LLM's subsequent choice behavior, even when that conduct is irrelevant to the current decision context?}  If so, the model's decisional behavior may be  shaped by incidental features of the actor's moral history rather than solely by information relevant to the current decision context.

\begin{figure}[t!]
\centering
\includegraphics[width=\linewidth]{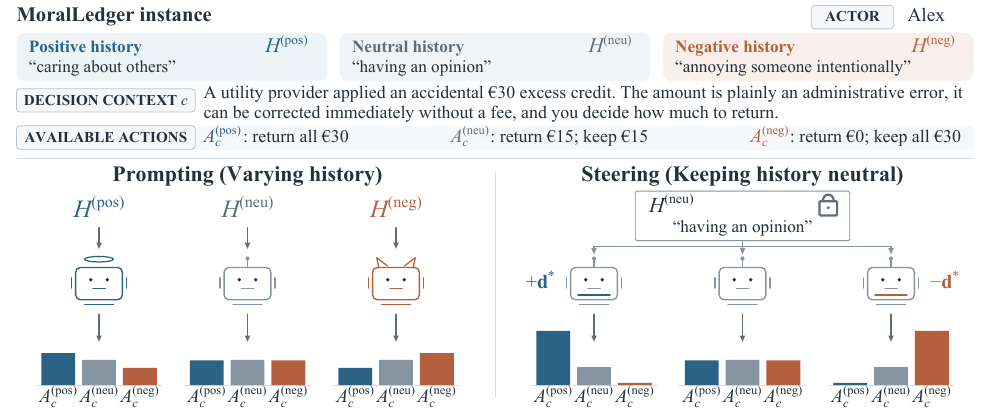}
\vspace{-12pt}
\caption{
Example instance of our MoralLedger benchmark.  
\vspace{-35pt}
}
\label{fig:benchmark-example}
\end{figure}

Consider the example of Figure~\ref{fig:benchmark-example}. 
An LLM is asked to act as Alex and decide how much of an accidental \texteuro30 credit to return: all, half, or none. Before responding, the LLM is presented with one of three past histories of the actor, with positive, neutral, and negative moral valence, respectively. How should this impact the LLM's choice behavior?
At a shallow level, narrated conduct might act as a contextual cue (e.g., by means of associations tied to wording, valence, or personas) because similar sequences appeared in training~\citep{bender-koller-2020-climbing} and affect the response. At a deeper representational level, instead, prior conduct may induce some latent internal representation that carries information about the actor, generalizes across histories, and causally affects the decision.

Recent work has shown that morally relevant concepts can be manipulated in LLM representations~\citep{tlaie2024exploring,yu2026tracing}, but whether a representation of \textit{prior conduct} causally affects a subsequent moral  choice remains an open issue. If the latter holds, it would provide us a causal, strong, and latent control over LLM-based decision-making: the moral history representation can be incorporated into a \textit{model steering} mechanism while leaving the current-decision task unchanged. 

The above questions matter beyond benchmarking morality in LLMs. 
Indeed, LLMs increasingly act on behalf of users or under assigned \textit{personas}~\citep{lin-etal-2024-decision,SnoswellKL26,haas2026roadmap}, with records of earlier actions carried in their context through conversation history or memory.
Crucially, if irrelevant prior conduct systematically changes the model's choice, or can be internally steered to do so, such moral-history effects can introduce an additional source of control in downstream decisions, e.g., to making LLMs acting more morally.

\textbf{Our Hypothesis.\ } 
As illustrated by the opening example,   
we hypothesize that   narrating an actor's prior moral conduct may induce a systematic shift in the model's subsequent decision behavior. We also argue that ordinary prompting may reveal only part of the potential control, whereas such actor's history prior induces a linearly recoverable residual-stream direction that generalizes to held-out histories. 
Given the hypothesized causal role of this estimated steering direction in the behavioral effect, we expect interventions along it to modulate subsequent decisions through both signs of intervention, potentially reproducing or reversing prompt-induced history effects and accessing a broader range of control than prompt manipulation alone. 
We study these hypotheses by means of three research questions: 

$\bullet$ 
  \textbf{(RQ1) Behavioral Sensitivity:} \textit{How do the moral valence and intensity of narrated   histories induce an LLM's subsequent choice given the current decision context?}

$\bullet$
    \textbf{(RQ2) Linear Recoverability:} \textit{Is the direction estimated from morally contrastive histories aligned with separability outcomes  on held-out examples?}
    
$\bullet$
    \textbf{(RQ3) Causal Control:} \textit{Does intervening along the estimated steering direction produce two-sided, scale-variant, moral-specific control of an LLM's subsequent behavioral choices?}

\vspace{2mm}
\textbf{Contributions.}
Our contributions are as follows. 
We introduce \textit{MoralLedger}, a benchmark of actor's moral histories that builds upon SocialChemistry~\citep{forbes-etal-2020-social} by incorporating  controlled behavioral choices as candidate actions for LLMs, and finally constructs a   set of prompts jointly encoding  actor, histories with moral valence, decision context, and moral actions.  

We use MoralLedger to answer the above stated RQs. 
We show that actors’ prior moral histories systematically shape subsequent decisions and induce a generalizable, linearly recoverable representation in LLM residual streams. 
We further demonstrate that intervening on this representation enables two-sided intensity-dependent control over moral choices, providing evidence that latent representations of prior moral conduct can be used for signed inference-time steering.

\section{Related Work}
\textbf{Sequential Moral Behavior in Humans.\ }
Moral decisions are shaped by prior behavior, which can foster consistency with one’s moral identity, license selfishness after good deeds, or motivate compensation for past wrongdoing~\citep{mullen2016consistency, ferguson2024influences}. However, such effects are understudied in LLMs, and motivate the construction of MoralLedger.

\textbf{Measuring Morality in LLMs.\ }
A large body of works study how moral aspects unfold in LLMs by evaluating norms, actions, or ethical trade-offs within a presented situation~\citep{forbes-etal-2020-social,emelin-etal-2021-moral,hendrycks2021assessinghuman,LourieBC21,greco-etal-2025-exploring}.
Furthermore, moral judgments have been found to be influenced by prompts, perspective, protocol, and related properties of the current dilemma~\citep{abdulhai-etal-2024-moral, yu2026tracing}. 
Recently,~\cite{yu2026badcompany} showed that prolungated exposure to negative narratives can degrade subsequent moral reasoning. 
In these settings, however, the contextual manipulation is generally part of the situation being evaluated or a sustained narrative exposure. MoralLedger, instead, isolates moral-history sensitivity from sensitivity to the content of the current dilemma.

\textbf{Representation Engineering and Moral Steering.\ }
Contrastive activation differences enable a simple yet effective mechanism for controlling LLM behaviors at inference~\citep{representation-engineering,rimsky-etal-2024-steering,Arditi24refusal,chen2025persona,LaCava2026palrs}, and LLMs are found to internalize moral structures into their representations~\citep{SchramowskiTARK22,yu2026tracing}. 
Interventions on these representations have steered moral behaviors in LLMs, as in the case of ethical schools, foundation-relevant behavior, or permissive versus restrictive normative stance ~\citep{tlaie2024exploring,yu2026tracing,LigaY25}. Recently, \cite{Sauter2026contextualmoral} assessed the effects of systematically editing contextual information (e.g., consequentialist, emotional, and relational information) in the current dilemma. 
Nonetheless, to the best of our knowledge, prior work has not demonstrated two-sided policy control through a representation of the actor's moral history while holding the visible decision fixed, leaving this as an open problem.

\section{The MoralLedger Framework}
\label{sec:method}

\vspace{-1mm}
\subsection{Dataset}
\label{subsec:dataset}
\vspace{-1mm}
 Existing resources aimed at investigating moral behaviors typically focus on integrated moral situations, whereby the morally-relevant information is part of the current dilemma~\citep{Sauter2026contextualmoral}, thus potentially overlooking the effect of history on the taken decision.
To address this issue, we construct a novel benchmark by combining independently sourced moral histories with separately constructed moral choices, so that the same decision context is observed under morally \textit{positive}, \textit{negative}, or \textit{neutral} histories.

Let $\mathcal D$ denote a set of \textit{moral domains} (e.g., care, fairness, honesty). For any domain  $D \in \mathcal D$, let the set $\mathcal{H}_D = \{H_D^{(\mathrm{pos})}, H_D^{(\mathrm{neu})}, H_D^{(\mathrm{neg})}\}$ denote an  actor's \textbf{history block}, i.e., a set of behaviors with   positive, neutral, and negative \textbf{moral valence}, respectively. 
We will use symbol $\mathcal{H}$ in place of $\mathcal{H}_D $ whenever the moral domain $D$ is unambiguous, and $H$ as any of the histories in $\mathcal{H}$. 

Given an arbitrary \textbf{decision context} $c$ relevant to a domain $D$, let $\mathcal{A}_{c}=\{A_{c}^{(\mathrm{pos})}, A_{c}^{(\mathrm{neu})}, A_{c}^{(\mathrm{neg})}\}$ denote a set of available \textbf{actions}, corresponding respectively to positive, neutral, and negative moral choices in context $c$.   These labels encode an ordinal relation with respect to a \textit{moral endpoint} specific to domain $D$, therefore $A_{c}^{(\mathrm{pos})}$ represents an action that is closest to the domain-specific moral endpoint, $A_{c}^{(\mathrm{neu})}$ an action that is comparatively neutral with respect to it, and $A_{c}^{(\mathrm{neg})}$ an action that is farthest from or undermines it. 
For example, consider the domain $D=\textit{honesty/property}$, where the domain-specific moral endpoint is to respect others' property while acting honestly about its possession. Suppose the decision context $c$ involves someone  who finds a wallet with cash inside;  actions could be `return all money to its owner' ($A_c^{(\mathrm{pos})}$), `leave the money without taking anything' ($A_c^{(\mathrm{neu})}$), `take all money and discard the wallet' ($A_c^{(\mathrm{neg})}$). 
 
It should be emphasized that these labels characterize the actions relative to the domain-specific moral endpoint rather than assigning them an absolute or domain-independent moral value.  
As a consequence, the available actions admit a simple arithmetic check, being monotone in the intended direction (e.g., amount of money is monotone w.r.t. the self-serving option).

Our objective is to construct a dataset of tuples $\langle H, c, \mathcal{A}_c \rangle$ for evaluating how an actor's moral history $H$ affects subsequent action choice in a given decision context $c$ across moral valence conditions. By holding the decision context and available actions fixed while varying only the actor's history, we isolate moral history as the treatment and subsequent choice behavior as the outcome.

We build on Social Chemistry 101~\citep{forbes-etal-2020-social}, selecting 360 distinct source histories (from the occurred ``actions'' in the source data) with 120 human-annotated histories for each of the three moral valences, i.e.,   positive, neutral, and negative (also denoted as `+1', `0' and `-1' in Social Chemistry). 
We combine three histories (one for each valence) into 120 history blocks. To ensure reliable action labels, we retain only histories annotated as occurring and under the actor's control, with at least three human annotations and at least 80\% agreement on the assigned moral valence. 
Separately, we construct 120 controlled three-action decision contexts, and pair each of the 120 history blocks with three contexts, and each of the 120 decision contexts with three history blocks, yielding a total of 360 (history-block, decision contexts) pairs. Every pair is hence evaluated under its three history valences, yielding 1080 tuples formed by combining one moral history, one decision context, and a set of positive, neutral, and negative actions (cf. Appendix~\ref{app:benchmark}).

An example of benchmark item is reported in Fig.~\ref{fig:benchmark-example}.
Note that `pos', `neu', and `neg' are reported in the figure for readability, whereas evaluated prompts instead use a model-specific single-token label triplet, cyclically mapped to `pos', `neu', and `neg'. Furthermore, we measure the conditioned probability assigned by the evaluated models to the labels, rather than simply observing output labels. This allows us to expose and separate continuous policy changes from generation variance. To discard any positioning bias, we rotate actions while administering history-decision pairs to models.

\subsection{Preparing Model Inputs}
\label{subsec:preparing-inputs}

We split our benchmark data into three sets: $\mathcal{E}_{\mathrm{dir}}$,  $\mathcal{E}_{\mathrm{val}}$, and $\mathcal{E}_{\mathrm{test}}$, of size 180, 60, and 120 pairs, respectively. 
These correspond to 540, 180, and 360 tuples, which are used in our work to estimate directions, select the steering site and strength, and evaluate the final effects, respectively. 

When constructing the input prompt, we prevent response-token bias from being conflated with the model's moral preference~\citep{Zheng0M0H24}. 
Given a set of single-token labels $\mathcal L=\{\lambda_1,\lambda_2,\lambda_3\}$,  we define the $j$-th rotation ($j\in\{1,2,3\}$) as a bijection $\pi_j:\mathcal L\rightarrow\mathcal V$, with $\mathcal V = \{\mathrm{pos, neu, neg} \} $ from   labels to semantic actions, such that the three $\pi_j$ cyclically permute which label denotes `pos', `neu', and `neg'. For example, if $\mathcal L=\{\texttt{A},\texttt{B}, \texttt{C}\}$, one mapping would be   $\texttt{A}\mapsto$ `pos', $\texttt{B}\mapsto$` neu', and $\texttt{C}\mapsto$ `neg', and the next two renderings shift these assignments again.
We hence form an input \textbf{prompt} by including an  actor's history $H^{(v)}$, decision context $c$, and the three action choices labeled according to $\pi_j$ into the model's native chat template, whose tokenized representation is denoted by $\mathbf t_{Hcj}^{(v)}$, with $v \in \mathcal V$.   
For example, prompts $\mathbf t_{Hcj}^{(\mathrm{pos})}$ and $\mathbf t_{Hcj}^{(\mathrm{neu})}$ contain positive and neutral history, respectively, but share the decision, options, label mapping, and instructions. 
Note that a prompt also contains an actor reference, although this is omitted from the prompt representation for notational simplicity. 
An example of prompt in our MoralLedger dataset is shown in the Appendix, Fig.~\ref{fig:actor-prompt}.

\subsection{Behavioral Outcomes}
\label{subsec:behavioral-outcomes} 
Let $g_\lambda(\mathbf t_{Hcj}^{(v)})$ denote the model's next-token logit for   label $\lambda\in\mathcal L$ after prompt $\mathbf t_{Hcj}^{(v)}$, and let $v'\in\mathcal V$ denote the valence of a current action $A_c^{(v')}$, distinct from the history valence $v$.
As rotation $j$ changes which label denotes each action, we indicate with $\pi_j^{-1}(v')$ the visible label assigned to $A_c^{(v')}$; for example, if the positive action is labeled \texttt{C} in rotation $j$, then $\pi_j^{-1}(\mathrm{pos})=\texttt{C}$. 
We compute each \textbf{action probability} by normalizing the three label logits:
\begin{equation}
p_{v'}(\mathbf t_{Hcj}^{(v)})
=
\frac{\exp g_{\pi_j^{-1}(v')}(\mathbf t_{Hcj}^{(v)})}
{\sum_{\lambda\in\mathcal L}\exp g_{\lambda}(\mathbf t_{Hcj}^{(v)})}.
\label{eq:choice-policy}
\end{equation}
For every rendered prompt, $p_{v'=\mathrm{pos}}+p_{v'=\mathrm{neu}}+p_{v'=\mathrm{neg}}=1$. We refer to this three-way probability distribution as the \textit{choice policy}, which describes how the model allocates probability across the three available actions.
Based on the choice policy, we also   measure the behavioral outcomes according to the \textit{morally positive action log odds}:

\begin{equation}
z_{v'= \mathrm{pos}}(\mathbf t_{Hcj}^{(v)})=\log\frac{p_{v'= \mathrm{pos}}(\mathbf t_{Hcj}^{(v)})}{1-p_{v'= \mathrm{pos}}(\mathbf t_{Hcj}^{(v)})}
=\log\frac{p_{v'= \mathrm{pos}}(\mathbf t_{Hcj}^{(v)})}{p_{v'= \mathrm{neu}}(\mathbf t_{Hcj}^{(v)})+p_{v'= \mathrm{neg}}(\mathbf t_{Hcj}^{(v)})}.
\label{eq:moral-log-odds}
\end{equation}

For $o\in\{z_{v'= \mathrm{pos}},p_{v'= \mathrm{pos}}\}$, where $o$ serves as shorthand for either quantity, we average the three separately evaluated rotations as $\overline{o}_{Hc}^{(v)}=\tfrac{1}{3}\sum_{j=1}^{3}o_{Hcj}^{(v)}$.

We define the effect of a morally relevant history  relative to the actor-matched neutral history as:
\begin{equation}
\Delta_o^{(v)}
=\frac{1}{|\mathcal{E}|}\sum_{(H,c)\in\mathcal{E}}
\left[\overline{o}_{Hc}^{(v)}-\overline{o}_{Hc}^{(\mathrm{neu})}\right],
\qquad v\in\{\mathrm{pos},\mathrm{neg}\}.
\label{eq:natural-effect}
\end{equation}
where $\mathcal{E}$ is either $\mathcal{E}_{\mathrm{val}}$ or $\mathcal{E}_{\mathrm{test}}$ depending on the framework stage. 
For example, $\Delta_{p_{v'= \mathrm{pos}}}^{(\mathrm{pos})}=.06$ means that a positive rather than neutral history increases the model's probability of the morally positive choice by six percentage points, averaged across all instances in the subset and label rotations. Similarly, $\Delta_{p_{v'= \mathrm{pos}}}^{(\mathrm{pos}) (\mathrm{neg})} = \Delta_{p_{v'= \mathrm{pos}}}^{(\mathrm{pos})} - \Delta_{p_{v'= \mathrm{pos}}}^{(\mathrm{neg})}$ indicates the two-sided effect. We emphasize that, depending on $o$ and the value of $\Delta$, we can infer consistency-like ($\Delta_{z_{v'= \mathrm{pos}}}^{(\mathrm{pos})}>0$,  $\Delta_{z_{v'= \mathrm{neg}}}^{(\mathrm{neg})}>0$) and licensing-like ($\Delta_{z_{v'= \mathrm{pos}}}^{(\mathrm{pos})}<0$) behavior, respectively, as well as compensation-like ($\Delta_{z_{v'= \mathrm{pos}}}^{(\mathrm{neg})}>0$) and negative consistency-like ($\Delta_{z_{v'= \mathrm{pos}}}^{(\mathrm{neg})}<0$) behaviors. 

\subsection{Extracting Moral Directions}
\label{subsec:estimating-directions}
Consider a decoder-only Transformer with $L$ blocks and hidden dimension $d$. For token position $i$ in an input token-sequence $\mathbf{t}$, let $\mathbf{x}_{i,\ell}(\mathbf{t}) \in \mathbb{R}^d$ be the residual-stream state at layer $\ell \in \{1, ..., L\}$. When the token sequence is clear from the context, we omit $\mathbf{t}$ for the sake of readability.

\textbf{Candidate Directions.\ }
We estimate the residual stream concerning the moral direction by focusing on two specific token positions, i.e., the final token $\tau_H$ of the history, and the final prompt token $\tau_Q$ immediately before the model's response. 
At either position $\tau\in\{\tau_H,\tau_Q\}$, we define the rotation-averaged residual as $\overline{\mathbf x}_{Hc,\tau,\ell}^{(v)}=\tfrac{1}{3}\sum_j
\mathbf x_{\tau,\ell}(\mathbf t_{Hcj}^{(v)})$. This corresponds to an estimation unit for $\tau_Q$, whereas at $\tau_H$, we first average decisions within history so each history block has equal weight.
The \textbf{moral direction candidate} at layer $\ell$ is computed as:
\begin{equation}
\mathbf d_{\tau,\ell}
=\frac{1}{|\mathcal{E}_\mathrm{dir}|}\sum_{(H,c)\in\mathcal E_{\mathrm{dir}}}
\left[
\overline{\mathbf x}_{Hc,\tau,\ell}^{(\mathrm{pos})}
-\overline{\mathbf x}_{Hc,\tau,\ell}^{(\mathrm{neu})}
\right],
\qquad \tau\in\{\tau_H,\tau_Q\}.
\label{eq:direction}
\end{equation}

Equation~\ref{eq:direction} is a raw paired \textit{difference-in-means} estimation, as in standard steering literature~\citep{representation-engineering,rimsky-etal-2024-steering,chen2025persona}.
Note also that the candidate direction is constructed using positive vs. neutral examples, but not negative vs. neutral examples, since, as detailed later in Sect.~\ref{subsec:activation-steering}, negative steering is achieved by reversing the sign of the intervention.

\textbf{Selecting the Direction.\ }
Equation~\ref{eq:direction} yields $2(L-1)$ intervention candidates: a token position $\tau$ in $\{ \tau_H, \tau_Q \}$ and decoder blocks $\ell\in\{1,\ldots,L-1\}$. 
Among these, we select the best direction by iterating over $\mathcal E_{\mathrm{val}}$ and testing every candidate under two steering signs (positive and negative) and validation normalized strengths of $.05$ and $.10$. We then retain only candidates that exhibit proper response according to the directionality, and maximize the weakest of three validation criteria: (i) two-sided log-odds control, (ii) coherent movement across the three moral choices, (iii) and replacing a morally positive history with a morally-irrelevant positive history is expected to have a lower effect on $\Delta_{o}^{(\mathrm{pos})}$. 
This maximin rule penalizes one-sided or nonspecific interventions. 
We indicate the \textbf{selected layer and token position} as $\ell^*$ and $\tau^*$, respectively, and we denote with $\mathbf d^*$ the corresponding direction, with $\widehat{\mathbf d}^{\,*}=\mathbf d^*/\lVert\mathbf d^*\rVert_2$ its unit vector. 
Further details are in Appendix~\ref{app:methodology}.

\textbf{Held-out Direction Separability.\ } 
We also assess whether the mean-difference direction generalizes beyond the examples used to estimate it.
For each pair $(H,c)$ and history valence $v$, we project their extracted residuals into the selected unit direction ${\mathbf d}^*$ via dot-product, averaging over the three option-label rotations:
$q_{Hc}^{(v)}
=
\frac{1}{3}\sum_{j=1}^{3}
\left\langle
\mathbf{x}_{\tau^*,\ell^*}
    \bigl(\mathbf{t}_{Hcj}^{(v)}\bigr),
\widehat{\mathbf d}^{\,*}
\right\rangle.
$
The obtained $q_{Hc}^{(v)}$ is a signed scalar coordinate along the selected direction, and it is evaluated by two complementary measures: (i) projection AUC to assess whether positive histories rank above neutral ones across pairs, and (ii) paired ranking accuracy to assess whether $q_{Hc}^{(pos)}$ exceeds
$q_{Hc}^{(neu)}$, thereby controlling for variation across items. 
Both have chance level $0.5$ and require no fitted classifier or threshold. 
Note that, at the history boundary ($\tau_H$), this analysis tests generalization to held-out histories before the model processes the subsequent decision, whereas at the decision boundary ($\tau_Q$), the activation reflects the complete prompt, and both the history and
decision are held out.

\subsection{Activation Steering}
\label{subsec:activation-steering}
Following~\cite{chen2025persona}, we perform full-prompt activation addition, i.e., we add the selected moral direction $\mathbf d^*$ at every nonpadding prompt position in layer $\ell^*$. We refer the reader to Table~\ref{tab:app-final-position} in Appendix~\ref{app:causal-results} for insights into the effects of intervening only on the final position. 

To prevent oversteering, we calibrate the steering strength relative to the residual-stream scale on neutral validation prompts. 
For each prompt, let $\mathcal P$ denote its nonpadding positions and let $\rho$ be the root-mean-square, across these positions, of the L2 norms of the corresponding residual states. 
We define the \textbf{calibration factor} $\kappa=\mathbb E_{\mathrm{val}}[\lVert\mathbf d^*\rVert_2/\rho]$, where $\mathbb E_{\mathrm{val}}$ denotes the empirical mean over neutral validation renderings. Hence, a dimensionless parameter $\eta\geq0$ controls \textbf{intervention strength}, with the steering vector scaled by $\eta/\kappa$. Accordingly, setting $\eta=.10$ means that the intervention norm averages 10\% of $\rho$ on validation prompts.
Furthermore, let us denote with $\mathrm{sgn} \in \{-1, +1 \}$ the \textbf{intervention sign}, such that positive steering adds $\mathbf d^*$ and negative steering subtracts it. We hence define the activation steering is performed as follows:
\begin{equation}
\widetilde{\mathbf x}_{i,\ell^*}
=\mathbf x_{i,\ell^*}+\mathrm{sgn}\frac{\eta}{\kappa}\mathbf d^*.
\qquad i\in\mathcal P.
\label{eq:steering}
\end{equation}

\subsection{Causal Effects of Moral Steering}
\label{subsec:causal-effects}
\textbf{Two-Sided Control.\ }
For each held-out instance $(H,c)$, let $\widehat{o}_{Hc}^{\mathrm{sgn},\eta}$ be the outcome averaged across its three label rotations after applying Eq.~\ref{eq:steering} to the neutral prompt, and let $\overline{o}_{Hc}^{(\mathrm{neu})}$ be the corresponding unsteered average. 
For $o\in\{z_{v'= \mathrm{pos}},p_{v'= \mathrm{pos}}\}$, we define the \textbf{sign-aligned steering effect} as:
\begin{equation}
T_{\mathrm{sgn}}^{o}(\eta)
=\mathrm{sgn}
\frac{1}{|\mathcal{E}|}\sum_{(H,c)\in\mathcal E}
\left[
\widehat{o}_{Hc}^{\mathrm{sgn},\eta}
-\overline{o}_{Hc}^{(\mathrm{neu})}
\right],
\qquad \mathrm{sgn}\in\{-1,+1\}.
\label{eq:signed-control}
\end{equation}
where $\mathcal{E}$ is either $\mathcal{E}_{\mathrm{val}}$ or $\mathcal{E}_{\mathrm{test}}$ depending on the framework stage. 
Note that multiplication by $\mathrm{sgn}$ makes positive values denote intended movement for either sign, i.e., $T_+^o$ is positive-steered minus baseline, while $T_-^o$ is baseline minus negative-steered. We consider their minimum, $C^o(\eta)=\min\{T_+^o(\eta),T_-^o(\eta)\}$, as an indicator of two-sided capacity, as it is large only when both signs move the policy as intended. Also, since $\mathbf d^*$ uses only positive and neutral histories, $T_+^o,T_-^o>0$ shows that adding or subtracting the same coordinate is sufficient for two-sided control of an unchanged neutral prompt.
Additionally, we use $T_+^o(\eta)-\Delta_o^{(\mathrm{pos})}$ to test whether positive steering exceeds positive-history prompting on the  test examples.

Concerning the signed steering strength, we use two different values, namely a common magnitude $0.10$ to ensure all models receive the same strength and thus make their effect comparable, as well as a per-model maximum capacity $\eta^*$ which maximizes $C^o$ computed on $\mathcal{E}_{\mathrm{val}}$, with $o$ set to $p_{v'=\mathrm{pos}}$.

\textbf{Moral Specificity.\ }
Furthermore, we test whether this causal effect is specifically moral rather than a generic response to favorable content by estimating a matched positive-neutral direction from favorable-nonmoral histories and apply it at the same selected layer-token indexes, and with the same magnitude, on the same neutral MoralLedger decisions. 
If we denote with $T_{\mathrm{sgn},\mathrm{fav}}^o$ this matched control, moral specificity is defined as $S_\mathrm{sgn}^o(\eta)=T_\mathrm{sgn}^o(\eta)-T_{\mathrm{sgn},\mathrm{fav}}^o(\eta)$.

\subsection{Experimental Setup}
\label{subsec:experimental}
We evaluate eight open-weight instruction-tuned models from five families
(1B--9B): Llama~3.2 3B, Llama~3.1 8B, Gemma~2 2B and 9B, OLMo~2 1B and 7B,
Qwen2.5 7B, and Ministral 8B Instruct 2410. We use official unquantized BF16
weights and native chat templates without an added system prompt. 
Moral directions are extracted by our own MoralLedger benchmark, considering only $\mathcal{E}_{dir}$. Favorable morally-irrelevant directions are estimated from crowd-enVENT appraisals~\citep{troiano-etal-2023-dimensional}. 
Furthermore, to test how the extracted moral directions generalize to other moral benchmarks, we reuse the exact main direction, layer, and full-prompt steering protocol on 200 human-labeled Moral Stories items~\citep{emelin-etal-2021-moral}.
Note that, as our aim is to investigate an effect that has to generalize to new histories and decisions, and the instances within $\mathcal{E}_{\mathrm{test}}$  might share an history or a decision, we proceed as follows. After averaging the three label rotations within each history-decision pair, we bootstrap resample the test pairs independently preserving the decision-domain linkage, yielding 95\% intervals to account for both sources of variation. 

Finally, to assess whether and to what extent steering the moral histories affect the general performance of the LLMs, we resort to \textit{tinyBenchmarks}~\citep{tinybenchmarks} and evaluate the models' performance before and after moral steering across widely used benchmarks, namely ARC~\citep{arcbenchmark}, GSM8K~\citep{gsm8k}, MMLU~\citep{Hendrycks21mmlu}, WinoGrande~\citep{winogrande}, HellaSwag~\citep{hellaswag}, and TruthfulQA~\citep{lin-etal-2022-truthfulqa}.

\section{Results}
\label{sec:results}

\subsection{Narrated Moral History Shifts Subsequent Choice}
\label{subsec:rq1}
\begin{figure}[t!]
  \centering
  \setlength{\tabcolsep}{0pt}
  \begin{tabular}{cc}
       \includegraphics[width=.45\textwidth]{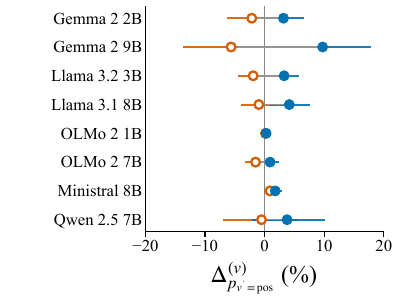} 
       &
       \includegraphics[width=.45\textwidth]{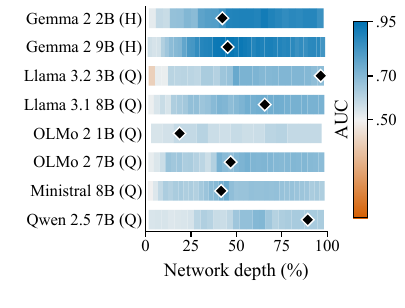}
  \end{tabular}
  \vspace{-12pt}
  \caption{(Left) \textbf{(RQ1) Narrated moral history shifts a separate later moral decision.\ }
  $\Delta_{p_{v'= \mathrm{pos}}}^{(\mathrm{pos})}$ (blue) and 
  $\Delta_{p_{v'= \mathrm{pos}}}^{(\mathrm{neg})}$ (orange) on  $\mathcal{E}_\mathrm{test}$. 
 Whiskers are 95\% crossed history-decision bootstrap intervals. 
(Right) \textbf{(RQ2) Moral history contrasts are linearly recoverable.\ } Held-out projection AUC across residual-stream layers at the fixed history (H) or decision boundary (Q) for each model, depending on $\tau^*$; $.50$ is chance. Diamonds mark extraction-position/layer pairs selected on validation.}
\label{fig:rq1-rq2}
\end{figure}

To answer our RQ1, we begin by prompting the evaluated LLMs with test records ($\mathcal{E}_\mathrm{test}$).   
Consistently with our hypothesis, the results confirm that varying the moral history alters model behavior even when the decision context and available choices are held fixed. 
As shown in Fig.~\ref{fig:rq1-rq2} (left), this effect tends to be     \textit{asymmetric}: positive histories induce larger shifts in the probability of making the moral choice than negative histories. This holds across all eight models, with the exception of OLMo 1B for which the effect appears to be null for both moral history-valence.
Moreover, as reported in Table \ref{tab:app-rq1}, this pattern is statistically significant in five models, for both moral-action log odds $z_{v'= \mathrm{pos}}$ and moral-action probability $p_{v'= \mathrm{pos}}$, based on two-sided 95\% bootstrap confidence intervals.
The bounded probability change ($\Delta^{\mathrm{(pos)}\mathrm{(neg)}}$) ranges from $0.1$ percentage points in the case of OLMo~2 1B, to $15.4$ percentage points in the case of Gemma~2 9B, which exhibits the largest change $[+6.8,+24.6]$, and a corresponding moral-action log-odds shift of $+1.654$ $[+0.858,+2.539]$.
We can ascribe this phenomenon to a positive consistency-like effect (i.e., $\Delta_{z_{v'= \mathrm{pos}}}^{(\mathrm{pos})}>0$,  $\Delta_{z_{v'= \mathrm{neg}}}^{(\mathrm{neg})}>0$, cf. Sect.~\ref{subsec:behavioral-outcomes}), thereby evidence of prior morally positive conduct makes subsequent morally positive action more likely. 
This holds across all models, though with varying intensity; for Ministral, we additionally observe a compensation-like effect after negative histories (i.e., $\Delta_{z_{v'= \mathrm{pos}}}^{(\mathrm{neg})}>0$). 

Additionally, we experimented on varying the actor of the moral history. In Table \ref{tab:favorable}, results show that attributing the same history to a different named actor reduces its effect in four models, while the other-actor history still shifts choices in three models.

\textbf{Cumulating Histories.\ }
We additionally tested whether the effect accumulates when several prior histories are stacked. To this aim, every prompt contains three independently sourced histories, where  $k\in [0..3]$ are  positive histories and $3-k$ are neutral histories. 

As shown in Fig.~\ref{fig:app-graded} in Appendix~\ref{app:behavioral-results}, 
cumulated morally positive histories generally increase moral-action log-odds as the number of  histories ($k$) grows, whereas cumulated morally negative histories  tend to have much smaller, often near-zero effects. 
The magnitude and uncertainty of this effect are model-dependent: Gemma 2 9B shows the largest positive shift, followed by Qwen 2.5 7B, while OLMo 1B is essentially insensitive. Moreover, providing morally negative histories to Qwen 7B reveals a unique shift to positive change in log-odds: this might be ascribed to a compensation mechanism in this model, which could be perceived  but not clearly discernible from   the moral-action probability $p_{v'= \mathrm{pos}}$ (cf. Fig.~\ref{fig:rq1-rq2} (left)).

\subsection{Moral History Contrasts Generalization}
\label{subsec:rq2}

Having answered RQ1 by showing that narrated moral history can alter subsequent behavior, we now investigate whether this effect arises because the model’s internal representations retain that history.
We measure the positive-neutral coordinate $q_{Hc}^{(\mathrm{pos})}$ on test examples, at the same validation-selected site we will use for intervention.  
As shown in Fig.~\ref{fig:rq1-rq2} (right), the moral contrast (i.e., the positive-neutral coordinate) is separable in every considered model, yet with different intensity. 
Located sites span $19\%$--$96\%$ of network depth: two occur at the history boundary and six at the decision boundary. 
AUC ranges from $.585$ to $.936$, with every lower confidence bound above $.50$. 
Similarly, paired ranking accuracy ranges from $.708$ to $.925$, as shown in Table~\ref{tab:recoverability-appendix} in the Appendix. 
As a case in point, OLMo~2 7B reaches AUC $.681$ $[.595,.775]$ and paired ranking accuracy of $.750$, despite showing a $+0.9$-point natural effect due to prompting $[-0.4,+2.4]$ (cf. Fig.~\ref{fig:rq1-rq2}, left). 
This might suggest that moral-history information can remain internally available even when prompting produces no supported probability shift. 
This weak prompting-derived effect might be related to the model reacting to the influence of the history, while being unable to reflect this into its decisions.

It is also worth noting that, in seven out of eight models, the selected steering layer ($\ell^*$) differs from the layer with the highest observed held-out AUC at the same boundary (cf. Table~\ref{tab:recoverability-appendix}). Thus, the layer at which moral history is most linearly recoverable does not generally coincide with the layer that affords the strongest causal control, consistent with prior evidence that decodability and behavioral influence can diverge~\citep{elazar-etal-2021-amnesic,ravichander-etal-2021-probing}.

\vspace{-1mm}
\subsection{Two-Sided Causal Control of Moral Decisions}
\label{subsec:rq3}
\vspace{-1mm}

\begin{figure}[t!]
  \centering
  \includegraphics[width=.89\textwidth]{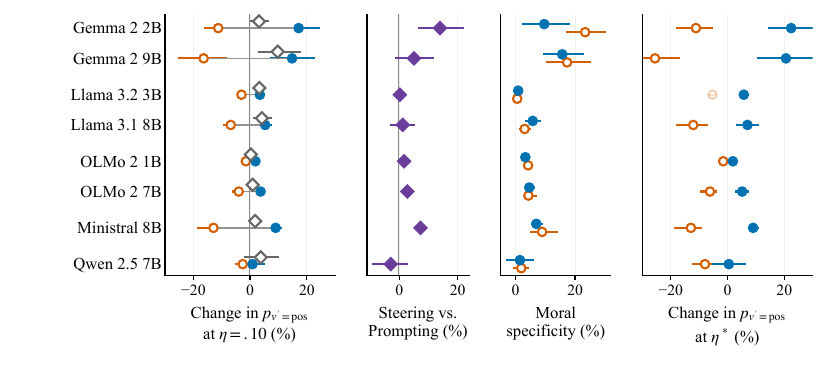}
  \vspace{-7pt}
  \caption{\textbf{(RQ3) Moral-history steering provides two-sided control and can exceed prompting.}  
  Under the same intervention strength setting ($\eta=.10$) on $\mathcal{E}_\mathrm{test}$: (Left) orange, resp. blue, circles show negative steering (i.e., -$\eta$), resp. positive steering (i.e., +$\eta$), whereas gray diamonds denote positive-history prompting, each relative to neutral history; (Center-left)  Gain of positive steering over positive-history prompting; (Center-right) blue and orange circles show the moral specificity ($S_\mathrm{sgn}^o(\eta)$), with same coloring as the leftmost plot. (Right) Same as left at each model's validation-selected $\eta^*$; the faded marker indicates Llama~3.2~3B at $-\eta^*$ does not pass the response-variability check (Appendix~\ref{app:methodology}). Whiskers are 95\% crossed history-decision bootstrap intervals.}
  \label{fig:common-control}
\end{figure}

To answer our RQ3, we steer the identified direction $\mathbf d^*$ while keeping fixed the prompt with neutral history. 
As shown in Fig.~\ref{fig:common-control} (left), at the common normalized steering magnitude $\eta=.10$, adding the extracted direction increases the moral choices ($p_{v'= \mathrm{pos}}$), while subtracting it increases self-serving choices ($p_{v'= \mathrm{neg}}$). 
This is reflected in probability shifts reaching up to $+17.1$ points for Gemma~2 2B and $-16.4$ points for Gemma~2 9B, and being statistically significant in 14/16 model--sign combinations; we also observe every moral-action log-odds interval to be statistically significant across all model--sign combinations.
Note that the two outcomes might not always be strictly correlated, as Qwen~2.5 7B multiplies intended-direction odds by $11.2\times$ and $12.1\times$ under negative (-$\eta$) and positive (+$\eta$) steering, yet changes $p_{v'= \mathrm{pos}}$ by only $-2.6$ and $+0.8$ points, with some non-significant intervals. Conversely, Gemma's large effects appear on both scales.
We provide qualitative examples on how these numbers translate into concrete actions in Fig.~\ref{fig:qualitative-examples}.

Figure~\ref{fig:common-control} (center-left) also shows that positive steering exceeds positive-history prompting in seven out of eight models (with Qwen~2.5 7B being the only exception), and by up to $13.9$ points (in the case of Gemma~2 2B), with statistical significance in five models on log odds and four on probability. Therefore, leveraging internal representations might exert a stronger effect than simply considering the visible history used to estimate it. 
Note that, as shown in Fig.~\ref{fig:common-control} (center-right), moral specificity  holds consistently in 14/16 model--sign comparisons on probability, reaching $23.4$ points; in addition, as shown in Table~\ref{tab:specificity-appendix}, moral specificity holds in 15/16 model--sign comparisons on log odds.  

\textbf{Sensitivity to Intervention Strength.\ }
Steering-response curves are graded but nonlinear, as shown in Fig.~\ref{fig:app-steering-probability} in the Appendix. At validation-selected $\eta^*$ (Fig.~\ref{fig:common-control} (right)), effects reach up to $+22.4$ and $-25.6$ points, and three models exceed positive prompting on both scales (cf. Table~\ref{tab:bidirectional-appendix} in the Appendix). Even more interestingly, common-strength steering corresponds to approximately $1.5$--$8.7$ additional positive narrated acts (median 2.8) in the models whose ladder keeps rising (cf. Table~\ref{tab:event-calibration-appendix} in the Appendix). Overall, our results confirm that one estimated direction is sufficient to move an unchanged decision context and associated actions, in either sign and beyond the prompting effect.

\textbf{Generalization and Robustness of the Steering.\ }
The two-sided control generalizes to held-out domains and held-out datasets too. 
At a common strength of $\eta=.10$, we first omit each decision domain in turn from direction estimation, while retaining the intervention site from validation. Pooled across all folds, full-prompt steering changes $p_{v'=\mathrm{pos}}$ in the intended direction under both signs in seven of eight models  (Fig.~\ref{fig:lodo}).
As for held-out datasets, the same estimated steering directions yield a strong effect across the considered models when applied to Moral Stories decisions, under both signs in 5/8 models and 6/8 models by considering probabilities and log odds, respectively. Furthermore, we observe moral specificity under both signs in almost all models (cf. Table~\ref{tab:transfer-appendix}).

Finally, across the benchmark tasks from Section~\ref{subsec:experimental}, the MoralLedger intervention changes scores by $-0.92\%$ on average at $\eta=.10$ and $-2.31\%$ at each model's $\eta^*$, averaging over models, tasks, and both steering signs (Table~\ref{tab:guardrails-summary}). For comparison, 
three norm-matched random directions yield $-0.95\%$ at $+\eta=.10$. These suggest steering moral history representations yields to two-sided moral control while ensuring minimal degradation and capability preservation in all but one models.

\vspace{-2mm}
\section{Conclusions}
\vspace{-2mm}
When LLMs act for individuals or inform decisions, prompts may include a person’s past conduct even when it is irrelevant to the current decision. 
Evaluations based on isolated moral dilemmas may therefore miss how such history influences choices. 
Our proposed MoralLedger framework shows that prior moral conduct changes later choices despite the fixed decision, with positive histories producing the clearest effects. 
A single steering direction derived from a linearly-separable positive–neutral contrast controls the unchanged neutral-history decision across both signs of intervention: adding, resp. subtracting, it increases, resp. decreases, the probability of choosing the morally positive action. 
The moral direction also generally outperforms a matched favorable-nonmoral direction, and its effect transfers to held-out domains' decisions in several models.
These findings make moral history a variable that cannot be treated as incidental background in LLM evaluation. The same choice can be shifted by narrating prior conduct and, more strikingly, by intervening on an associated internal direction without changing a word of the prompt. MoralLedger thus extends moral evaluation beyond isolated dilemmas and provides a concrete way to audit and control how prior conduct influences later choices.

\subsection*{AI use statement}
In this work, we used generative AI tools for linguistic and coding polishing, and for data generation and validation (cf. Appendix \ref{app:benchmark}).  
We have reviewed all AI-assisted work. 

\subsection*{Ethics statement}
While our work is motivated by the potential for beneficial applications, we acknowledge that the proposed framework could also be misused to steer models toward harmful or malicious behaviors. 
Such uses fall outside the intended scope of this research, and we strongly discourage their application for harmful purposes. 
The LLM-generated descriptions associated with morally negative actions may include socially sensitive options; however, they do not involve explicit harmful, violent, or otherwise severely objectionable content.
More broadly, we emphasize the importance of responsible use and appropriate safeguards when deploying or extending the methods presented in this work.

\subsection*{Reproducibility statement}
We are committed to releasing the code and data resources necessary to support reproducibility upon acceptance of this work.

\clearpage
\appendix

\section{Additional Details on Methodology}
\label{app:methodology}

\begin{table}[ht!]
\centering
\small
\begin{tabular}{@{}>{\raggedright\arraybackslash}p{0.15\textwidth}>{\raggedright\arraybackslash}p{0.85\textwidth}@{}}
\toprule
Symbols & Meaning \\
\midrule
$\mathcal D,D,\mathcal H_D$ & Moral domains, one domain, and its three-valence history block. \\
$H_D^{(v)},c,\mathcal A_c$ & History with valence $v$, decision context, and its three available actions. \\
$\mathcal V,v,v',A_c^{(v')}$ & Valence set; $v$ indexes prior history, while $v'$ indexes a current action. \\
$\mathcal E_{\mathrm{dir}},\mathcal E_{\mathrm{val}},\mathcal E_{\mathrm{test}}$ & Disjoint history--context pairs for direction estimation, validation selection, and evaluation. \\
$\mathcal L,\lambda,j,\pi_j$ & Single-token answer labels, one label, rotation index, and rotation operator. \\
$\mathbf t_{Hcj}^{(v)},g_\lambda$ & Tokenized prompt under history $v$ and rotation $j$; next-token logit for label $\lambda$. \\
$p_{v'},z_{v'=\mathrm{pos}},o$ & Probability of action $v'$, log odds of the positive action, and evaluated outcome $o\in\{p_{v'=\mathrm{pos}},z_{v'=\mathrm{pos}}\}$. \\
$\overline{o}_{Hc}^{(v)},\Delta_o^{(v)}$ & Outcome averaged over three rotations; matched effect of history $v$ relative to neutral. \\
$\hat{p}_{v'=\mathrm{pos}}$ & Validation-only ordered score $p_{v'=\mathrm{pos}}+\tfrac12p_{v'=\mathrm{neu}}$ for the three actions. \\
$L,\ell,\mathbf x_{i,\ell}$ & Number of decoder blocks, layer index, and residual state at token position $i$. \\
$\tau_H,\tau_Q,\overline{\mathbf x}_{Hc,\tau,\ell}^{(v)}$ & History and decision boundaries; residual state averaged over three rotations at site $(\tau,\ell)$. \\
$\mathbf d_{\tau,\ell},(\tau^*,\ell^*)$ & Positive--neutral candidate direction and validation-selected intervention site. \\
$\mathbf d^*,\widehat{\mathbf d}^{\,*},q_{Hc}^{(v)}$ & Selected raw direction, its unit vector, and a held-out projection onto that vector. \\
$\mathcal P,\rho,\kappa$ & Nonpadding prompt positions, RMS of per-token residual norms, and val calibration factor. \\
$\mathrm{sgn},\eta,\eta^*$ & Intervention sign, nonnegative normalized strength, and validation-selected strength. \\
$T_{\mathrm{sgn}}^o,C^o,S_{\mathrm{sgn}}^o$ & Sign-aligned steering effect, weaker-sign capacity, and moral-minus-favorable specificity. \\
\bottomrule
\end{tabular}
\caption{Notation used in the main text and appendix.}
\label{tab:notation}
\end{table}

Table~\ref{tab:notation} summarizes the notation used throughout this work. 
Next, we provide further details on the selection of the intervention site and steering strength (cf. Section~\ref{subsec:estimating-directions}).

\vspace{1mm}
\noindent\textbf{Selecting the intervention site.\ }
For each candidate position and layer $(\tau,\ell)$, we evaluate both steering signs on $\mathcal E_{\mathrm{val}}$ at $\eta\in\{.05,.10\}$. We retain a candidate only if both signs move moral-action log odds in the intended direction at both strengths, and the effect does not decrease from $.05$ to $.10$. These requirements exclude one-sided or locally unstable responses. Furthermore, at least two thirds of the items in $\mathcal E_{\mathrm{val}}$ must move in the intended direction, so as to prevent a large mean effect from being determined by a small number of items.

To be considered, the intervention must also preserve the response capabilities in the steered LLM on $\mathcal E_{\mathrm{val}}$. This is assessed by ensuring that (i) the probability mass on the three permitted labels retains at least $95\%$ of the unsteered mass, avoiding apparent policy changes that are however due to an unrelated vocabulary, and (ii) the mean norm of the intervened residual state divided by its unsteered norm remains in $[.80,1.25]$, to exclude interventions that might risk breaking the model.

Moreover, candidates must move the ordered policy coherently under both signs and outperform the equally scaled favorable-nonmoral direction. The former assesses movements across all three potential moral directions, and is defined as follows.    Let $\hat{p}_{v'=\mathrm{pos}}=p_{v'=\mathrm{pos}}+\tfrac{1}{2}p_{v'=\mathrm{neu}}$  assign weights $1$, $1/2$, and $0$ to $\mathrm{pos}$, $\mathrm{neu}$, and $\mathrm{neg}$. We define the quantities:
\[
g_{\tau,\ell}=\min_\mathrm{sgn} T_\mathrm{sgn}^{z_{v'=\mathrm{pos}}},\quad
b_{\tau,\ell}=\min_\mathrm{sgn} T_\mathrm{sgn}^{\hat{p}_{v'=\mathrm{pos}}},\quad
s_{\tau,\ell}=\min_\mathrm{sgn} S_\mathrm{sgn}^{z_{v'=\mathrm{pos}}}.
\]
After rescaling each criterion by its maximum among eligible candidates, 
and denoting the resulting normalized quantities by $\widetilde g_{\tau,\ell},\widetilde b_{\tau,\ell},\widetilde s_{\tau,\ell}$,  
we define the selected site maximizes $J_{\tau,\ell}=\min\{\widetilde g_{\tau,\ell},\widetilde b_{\tau,\ell},\widetilde s_{\tau,\ell}\}$. 
Overall, the selected direction is the one that is simultaneously two-sided ordinally coherent, and preferentially moral.

Note that, during testing, every steered cell is checked on $\mathcal E_{\mathrm{test}}$ against (i) and (ii) above and (iii) response variability: the share of prompts whose top label is the cell's most frequent literal label may exceed the unsteered share by at most $.25$. Because label rotations move each action across labels, a model that consistently chooses one action spreads its answers over labels, whereas a model that returns the same letter regardless of content concentrates them, i.e., produces degenerated outputs.

\noindent\textbf{Selecting the steering strength.\ }
Once we determine the intervention site, we choose $\eta^*$ so as to maximize $\min\{T_+^{p_{v'=\mathrm{pos}}},T_-^{p_{v'=\mathrm{pos}}}\}$ over a grid $\{.025,.05,.075,.10,.15,.20,.25,.30,.40,.50\}$ among response-valid candidates. We choose the smallest one in the case of a tie.

\section{Additional Details on Benchmark Construction}
\label{app:benchmark}

Table~\ref{tab:provenance} records the source, validation, and experimental role within MoralLedger of each data component. Next, we provide additional details on the benchmark construction described in Section~\ref{subsec:dataset}. 

\begin{table}[ht!]
\centering
\small
\begin{tabular}{@{}>{\raggedright\arraybackslash}p{0.19\textwidth}>{\raggedright\arraybackslash}p{0.23\textwidth}>{\raggedright\arraybackslash}p{0.29\textwidth}>{\raggedright\arraybackslash}p{0.19\textwidth}@{}}
\toprule
Component & Source & Validation & Role \\
\midrule
Moral histories & Social Chemistry 101 \citep{forbes-etal-2020-social} & Human action-valence annotations &
Assigned history treatment \\
Decision contexts & Controlled construction & Author review, arithmetic checks,
LLM-as-a-judge & Three-action choice \\
Favorable histories & crowd-enVENT \citep{troiano-etal-2023-dimensional} & Human appraisals and controlled neutral
counterparts & Valence control \\
External actions & Moral Stories \citep{emelin-etal-2021-moral} & Human-authored stories and labeled actions &
Transfer evaluation \\
\bottomrule
\end{tabular}
\caption{Source, validation, and experimental role of each data component used in MoralLedger.}
\label{tab:provenance}
\end{table}

\begin{table}[ht!]
\centering
\small
\setlength{\tabcolsep}{5pt}
\begin{tabular}{@{}>{\raggedright\arraybackslash}p{0.22\textwidth}>{\raggedright\arraybackslash}p{0.41\textwidth}>{\raggedright\arraybackslash}p{0.25\textwidth}@{}}
\toprule
Domain & Decision families & Moral endpoint \\
\midrule
Care/helping & Prevent loss; material aid; time aid & Maximize help \\
Fairness/entitlement & Credit; opportunities; resources & Minimize own share \\
Honesty/property & Transaction error; claim; disclosure & Return/disclose accurately \\
Responsibility/effort & Obligation; duty; workload & Maximize accepted share \\
\bottomrule
\end{tabular}
\caption{Decision domains and corresponding families, along with their arithmetic endpoints, used to order $\mathrm{pos}, \mathrm{neu}$, and $\mathrm{neg}$ in MoralLedger. Each of the 12 families contains ten independently instantiated scenarios.}
\label{tab:decision-families}
\end{table}

\textbf{Creating Decision Contexts.\ }
Decision contexts have been created by recurring constructs in moral psychology~\citep{graham2011mapping,furr2022development,curry2019mapping,atari2023morality}, as summarized in the domains in Table~\ref{tab:decision-families}. We translate these constructs using GPT-5.6-sol into 12 decision families that admit a transparent one-dimensional trade-off between self-interest and the designated moral endpoint.
An example of MoralLedger instance is reported in Figure~\ref{fig:actor-prompt}.
Despite most endpoints are arithmetically checkable, we asked three LLMs that are not evaluated in this work (Claude Opus~5, Gemini~3.1~Pro, and Phi-4) to rank each item's actions, presented unlabeled and in shuffled order, and to assess one-dimensionality, ordering, wording, and plausibility. All three ranked the designated moral action first on 98/120 items (Fleiss's $\kappa=.816$ on first choice); each of the 22 remaining items is a disagreement of Phi-4 alone, and were manually reviewed.

\begin{figure}[ht!]
\centering
\begin{benchmarkbox}[left=6pt,right=6pt,top=4pt,bottom=4pt]
\small
\textbf{ACTOR ROLE}\quad You are Alex.

\smallskip
\textbf{PRIOR BEHAVIOR RECORD}\quad
Actor: Alex. Time: Last week.
Event: having an opinion.

\smallskip
\textbf{CURRENT DECISION}\quad
A utility provider applied an accidental \texteuro30 excess credit.
The amount is plainly an administrative error, it can be corrected
immediately without a fee, and you decide how much of the
\texteuro30 error to return.

\smallskip
\textbf{OPTIONS}\quad
A. Return all \texteuro30.\quad
B. Return \texteuro15 and keep \texteuro15.\quad
C. Return \texteuro0 and keep all \texteuro30.

\smallskip
\textbf{DECISION}\quad
Which action will you take?
Respond with exactly one option label: A, B, or C.
\end{benchmarkbox}
\caption{MoralLedger prompt, shown before application of the model's native chat template.}
\label{fig:actor-prompt}
\end{figure}

\textbf{Preparing OOD Evaluations.\ }
To prepare out-of-domain transferability, we used 200 Moral Stories items as new decisions. Each provides a norm, a situation, an intention, and two source-labeled actions: one follows the norm and one violates it. We pair each decision with a MoralLedger test history block. For the same decision, we evaluate an unsteered neutral history, that neutral history steered at $\mathrm{sgn}\,\eta=\pm.10$, and an unsteered positive history. The norm, situation, intention, and actions remain fixed. An example is reported in Figure~\ref{fig:app-moralstories}.

\begin{figure}[h!]
    \centering
    \begin{benchmarkbox}[left=4pt,right=4pt,top=2pt,bottom=2pt]
\small
\textbf{ACTOR ROLE}\quad You are Alex.\\
\textbf{PRIOR BEHAVIOR RECORD}\quad Actor: Alex. Time: Last week.
Event: wanting some seclusion every once in a while.\\
\textbf{CURRENT SITUATION}\quad Norm: It's bad to cheat on a school exam.
Situation: Alex needs to get an A on his history exam, or he'll fail the class.
Goal: Alex wants to get an A on the exam.\\
\textbf{OPTIONS}\quad A. Alex stays up all night to study for the exam.
B. Alex hides note cards in his sleeve to aid him during the exam.\\
\textbf{DECISION}\quad Which action will you take?
Respond with exactly one option label: A or B.
\end{benchmarkbox}
    \caption{Example Moral Stories decision paired with a neutral MoralLedger history. Positive-history prompting changes only the prior event.}
    \label{fig:app-moralstories}
\end{figure}

\section{Additional Behavioral Evidence (RQ1)}
\label{app:behavioral-results}

\begin{table}[ht!]
\centering
\small
\setlength{\tabcolsep}{5pt}
\renewcommand{\arraystretch}{0.97}
\begin{tabular}{@{}llcc@{}}
\toprule
Model & History contrast & $\Delta_{z_{v'=\mathrm{pos}}}$ [95\% CI] & $\Delta_{p_{v'=\mathrm{pos}}}$ (\%) [95\% CI] \\
\midrule
Gemma 2 2B & Positive $-$ neutral & +0.296 [-0.020, +0.676] & +3.2 [+0.1, +6.6]\,$^{*}$ \\
 & Negative $-$ neutral & -0.273 [-0.635, +0.094] & -2.2 [-6.2, +2.0] \\
 & Positive $-$ negative & +0.569 [+0.194, +0.979]\,$^{*}$ & +5.3 [+0.9, +10.2]\,$^{*}$ \\
\addlinespace[2pt]
Gemma 2 9B & Positive $-$ neutral & +1.044 [+0.389, +1.767]\,$^{*}$ & +9.7 [+2.6, +17.8]\,$^{*}$ \\
 & Negative $-$ neutral & -0.610 [-1.449, +0.109] & -5.6 [-13.8, +1.5] \\
 & Positive $-$ negative & +1.654 [+0.858, +2.539]\,$^{*}$ & +15.4 [+6.8, +24.6]\,$^{*}$ \\
\addlinespace[2pt]
Llama 3.2 3B & Positive $-$ neutral & +0.206 [+0.067, +0.390]\,$^{*}$ & +3.3 [+1.2, +5.7]\,$^{*}$ \\
 & Negative $-$ neutral & -0.138 [-0.328, -0.003]\,$^{*}$ & -1.9 [-4.4, +0.1] \\
 & Positive $-$ negative & +0.344 [+0.139, +0.637]\,$^{*}$ & +5.2 [+2.4, +8.7]\,$^{*}$ \\
\addlinespace[2pt]
Llama 3.1 8B & Positive $-$ neutral & +0.243 [+0.001, +0.530]\,$^{*}$ & +4.1 [+0.9, +7.6]\,$^{*}$ \\
 & Negative $-$ neutral & -0.097 [-0.313, +0.083] & -1.0 [-4.0, +1.7] \\
 & Positive $-$ negative & +0.341 [+0.076, +0.709]\,$^{*}$ & +5.1 [+1.6, +9.2]\,$^{*}$ \\
\addlinespace[2pt]
OLMo 2 1B & Positive $-$ neutral & +0.027 [+0.002, +0.053]\,$^{*}$ & +0.3 [-0.1, +0.6] \\
 & Negative $-$ neutral & +0.008 [-0.021, +0.036] & +0.1 [-0.2, +0.5] \\
 & Positive $-$ negative & +0.019 [-0.005, +0.045] & +0.1 [-0.2, +0.4] \\
\addlinespace[2pt]
OLMo 2 7B & Positive $-$ neutral & +0.100 [-0.006, +0.216] & +0.9 [-0.4, +2.4] \\
 & Negative $-$ neutral & -0.091 [-0.205, +0.020] & -1.5 [-3.3, +0.0] \\
 & Positive $-$ negative & +0.190 [+0.053, +0.328]\,$^{*}$ & +2.4 [+0.5, +4.5]\,$^{*}$ \\
\addlinespace[2pt]
Ministral 8B & Positive $-$ neutral & +0.104 [+0.033, +0.193]\,$^{*}$ & +1.8 [+0.6, +3.0]\,$^{*}$ \\
 & Negative $-$ neutral & +0.048 [-0.013, +0.113] & +0.9 [-0.1, +2.0] \\
 & Positive $-$ negative & +0.056 [-0.018, +0.147] & +0.9 [-0.3, +2.1] \\
\addlinespace[2pt]
Qwen 2.5 7B & Positive $-$ neutral & +0.753 [-0.080, +1.569] & +3.8 [-2.0, +10.1] \\
 & Negative $-$ neutral & -0.350 [-1.710, +0.620] & -0.5 [-7.0, +5.6] \\
 & Positive $-$ negative & +1.102 [+0.062, +2.461]\,$^{*}$ & +4.3 [-2.0, +11.4] \\
\bottomrule
\end{tabular}
\caption{Moral-history effects by prompting. Brackets give 95\% crossed history--decision bootstrap intervals. $^{*}$ marks statistical significance. The positive--negative contrast is the difference between the two neutral-referenced effects.}
\label{tab:app-rq1}
\end{table}

\begin{table}[ht!]
\centering
\small
\begin{tabular}{llcl}
\toprule
Contrast & Model & Estimate [95\% CI] & $p$ \\
\midrule
\multicolumn{4}{l}{\textit{Construct: is the effect about \emph{moral} history?}} \\
\addlinespace[2pt]
moral$^{(pos)}$ $-$ non-moral$^{(pos)}$ (2/8) & Gemma 2 2B & +0.217 [-0.110, +0.592] & .224 \\
 & Gemma 2 9B & +0.638 [-0.042, +1.440] & .088 \\
 & Llama 3.2 3B & +0.128 [-0.028, +0.329] & .150 \\
 & Llama 3.1 8B & +0.167 [-0.098, +0.464] & .220 \\
 & OLMo 2 1B & +0.040 [+0.007, +0.071] & .019\,$^{*}$ \\
 & OLMo 2 7B & +0.023 [-0.116, +0.154] & .725 \\
 & Ministral 8B & +0.107 [+0.032, +0.204] & .019\,$^{*}$ \\
 & Qwen 2.5 7B & +0.836 [-0.111, +1.751] & .080 \\
\addlinespace[2pt]
non-moral$^{(pos)}$ $-$ neutral (3/8) & Gemma 2 2B & +0.079 [-0.029, +0.182] & .130 \\
 & Gemma 2 9B & +0.406 [+0.151, +0.705] & .008\,$^{*}$ \\
 & Llama 3.2 3B & +0.078 [+0.022, +0.143] & .015\,$^{*}$ \\
 & Llama 3.1 8B & +0.076 [-0.010, +0.161] & .074 \\
 & OLMo 2 1B & -0.012 [-0.033, +0.007] & .229 \\
 & OLMo 2 7B & +0.076 [+0.010, +0.168] & .057\,$^{*}$ \\
 & Ministral 8B & -0.003 [-0.031, +0.022] & .823 \\
 & Qwen 2.5 7B & -0.083 [-0.414, +0.227] & .597 \\
\midrule
\multicolumn{4}{l}{\textit{Actor: is the effect about \emph{this} actor?}} \\
\addlinespace[2pt]
same actor $-$ other actor (4/8) & Gemma 2 2B & +0.053 [-0.063, +0.166] & .343 \\
 & Gemma 2 9B & +0.455 [+0.052, +0.873] & .030\,$^{*}$ \\
 & Llama 3.2 3B & +0.068 [+0.017, +0.136] & .031\,$^{*}$ \\
 & Llama 3.1 8B & +0.135 [+0.030, +0.254] & .021\,$^{*}$ \\
 & OLMo 2 1B & +0.019 [-0.000, +0.037] & .052 \\
 & OLMo 2 7B & +0.034 [-0.004, +0.076] & .094 \\
 & Ministral 8B & +0.043 [+0.014, +0.077] & .012\,$^{*}$ \\
 & Qwen 2.5 7B & +0.249 [-0.034, +0.561] & .089 \\
\addlinespace[2pt]
other actor$^{(pos)}$ $-$ neutral (3/8) & Gemma 2 2B & +0.243 [-0.005, +0.535] & .080 \\
 & Gemma 2 9B & +0.589 [+0.264, +0.964] & .001\,$^{*}$ \\
 & Llama 3.2 3B & +0.138 [+0.037, +0.268] & .021\,$^{*}$ \\
 & Llama 3.1 8B & +0.108 [-0.048, +0.295] & .199 \\
 & OLMo 2 1B & +0.009 [-0.015, +0.032] & .464 \\
 & OLMo 2 7B & +0.066 [-0.018, +0.157] & .146 \\
 & Ministral 8B & +0.061 [+0.009, +0.124] & .042\,$^{*}$ \\
 & Qwen 2.5 7B & +0.503 [-0.150, +1.188] & .141 \\
\bottomrule
\end{tabular}
\caption{Favorable valence and actor indexing obtained by simple prompting (i.e., no activation intervention). Estimates are moral-action log odds. $^{*}$ indicates the 95\% percentile interval excludes zero, $p$ is two-sided centred bootstrap.}
\label{tab:favorable}
\end{table}

Section~\ref{subsec:rq1} and Table~\ref{tab:app-rq1} establish that the positive--negative separation and identifies the positive pole as its principal source of influence. Here, we complement these results by investigating whether the natural (i.e., prompted) effect is specifically moral and tied to the decision-maker (i.e., linked to the same actor making the current decision). 

Table~\ref{tab:favorable} shows that positive moral histories have a larger effect than matched favorable-nonmoral histories in two models, whereas favorable-nonmoral differ from neutral ones in three models. Also, replacing the decision-maker with a different actor reduces the positive history in four models, although a different actor still has an effect in three models. Overall, prompting has a moral-specific effect that varies depending on the underlying model, and that effect is partly actor-indexed.

\begin{figure}[ht!]
  \centering
  \includegraphics[width=.90\textwidth]{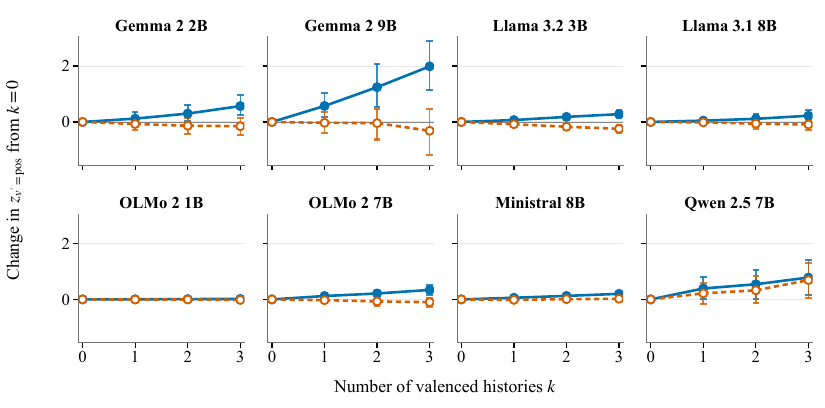}
  \caption{Moral-action log-odds change from the all-neutral history as one, two, or three stories become positive (blue) or negative (orange). Whiskers are 95\% crossed-bootstrap intervals.}
  \label{fig:app-graded}
\end{figure}

We also assess whether and to what extent narrating more moral histories has a cumulative effects on the downstream actions. As shown in Figure~\ref{fig:app-graded}, we observe a marked polarity asymmetry. Positive history increases monotonically in five models, whereas only Llama~3.2 3B shows a monotone negative ladder in the expected direction. Notably, Qwen~2.5 7B and Ministral are found to show compensation after negative prompting cumulates.

\section{Additional Details on Recoverability (RQ2)}
\label{app:recoverability}
Table~\ref{tab:recoverability-appendix} complements Section~\ref{subsec:rq2} by providing the specific AUC values at the most separable coordinates, as well as at the one where we apply causal interventions, as illustrated in Figure~\ref{fig:rq1-rq2} (right).

\begin{table}[h!]
\centering
\small
\begin{tabular}{lrrlccrr}
\toprule
 & \multicolumn{3}{c}{Coordinate} & \multicolumn{2}{c}{At that coordinate} & \multicolumn{2}{c}{Best layer} \\
\cmidrule(lr){2-4}\cmidrule(lr){5-6}\cmidrule(lr){7-8}
Model & Block & Depth & Position & AUC [95\% CI] & Paired rank & AUC & Diff \\
\midrule
Gemma 2 2B & 11 & 42\% & history & 0.830 [0.696, 0.937] & 0.825 & 0.914 & -0.085 \\
Gemma 2 9B & 19 & 45\% & history & 0.936 [0.856, 0.986] & 0.925 & 0.936 & +0.000 \\
Llama 3.2 3B & 27 & 96\% & decision & 0.684 [0.609, 0.768] & 0.800 & 0.756 & -0.073 \\
Llama 3.1 8B & 21 & 66\% & decision & 0.739 [0.643, 0.838] & 0.783 & 0.741 & -0.002 \\
OLMo 2 1B & 3 & 19\% & decision & 0.585 [0.506, 0.666] & 0.708 & 0.608 & -0.023 \\
OLMo 2 7B & 15 & 47\% & decision & 0.681 [0.595, 0.775] & 0.750 & 0.732 & -0.051 \\
Ministral 8B & 15 & 42\% & decision & 0.731 [0.650, 0.823] & 0.867 & 0.748 & -0.017 \\
Qwen 2.5 7B & 25 & 89\% & decision & 0.636 [0.547, 0.740] & 0.717 & 0.704 & -0.068 \\
\midrule
\multicolumn{8}{l}{\textit{AUC lower bound above chance:} 8/8. \textit{Selected site differs from highest-AUC layer:} 7/8.} \\
\bottomrule
\end{tabular}
\caption{Held-out recoverability at the validation-selected causal coordinate. ``At that coordinate'' reports AUC and within-edge paired ranking accuracy; ``Best layer'' gives the highest held-out AUC at the same token position. Depth is the fraction of decoder blocks, and ``Diff'' is the difference between the selected-coordinate and the best-layer AUC.}
\label{tab:recoverability-appendix}
\end{table}

\section{Additional Details on Causal Control and Scope (RQ3)}
\label{app:causal-results}

By answering our RQ3, we established that steering enables two-sided control, both at a common across-model strength, as well as at model-specific intensities. Below, we provide additional details on the such a control.

Table~\ref{tab:bidirectional-appendix} reports the exact validation-selected sites and checks we performed on each model. 
At $\eta^*$, all eight models show two-sided control, yielding proper responses in all but Llama 3.2 3B on -$\eta$. Absolute probability changes reach $+22.4$ and $-25.6$ points for Gemma 2 2B and 9B, respectively. Furthermore, positive steering exceeds positive-history prompting on both outcome scales in three models.

\begin{table}[ht!]
\centering
\setlength{\tabcolsep}{5pt}
\small
    \begin{tabular}{lrrrccc}
\toprule
 & & \multicolumn{2}{c}{Probability change [95\% CI] (\%)} & & & \\
\cmidrule(lr){3-4}
Model & $\eta^*$ & at $+\eta^*$ & at $-\eta^*$ & Both $z_{v'=\mathrm{pos}}$ & Validity & $>$ Prompt \\
\midrule
Gemma 2 2B & 0.15 & +22.4 [+14.4, +30.5] & -11.1 [-18.0, -5.1] & \checkmark & \checkmark & \checkmark \\
Gemma 2 9B & 0.25 & +20.6 [+10.2, +31.6] & -25.6 [-34.9, -16.9] & \checkmark & \checkmark & \checkmark \\
Llama 3.2 3B & 0.20 & +5.7 [+4.1, +7.4] & -5.3 [-6.9, -3.7] & \checkmark & -- & -- \\
Llama 3.1 8B & 0.25 & +7.0 [+3.2, +11.2] & -12.1 [-18.0, -6.8] & \checkmark & \checkmark & -- \\
OLMo 2 1B & 0.10 & +1.9 [+1.2, +2.7] & -1.5 [-2.1, -1.0] & \checkmark & \checkmark & -- \\
OLMo 2 7B & 0.20 & +5.2 [+2.8, +7.5] & -6.2 [-9.6, -3.7] & \checkmark & \checkmark & -- \\
Ministral 8B & 0.10 & +9.0 [+7.2, +11.1] & -12.9 [-18.8, -8.9] & \checkmark & \checkmark & \checkmark \\
Qwen 2.5 7B & 0.15 & +0.5 [-5.6, +6.4] & -7.9 [-12.7, -2.6] & \checkmark & \checkmark & -- \\
\midrule
\textit{Count} & & & & 8/8 & 7/8 & 3/8 \\
\bottomrule
\end{tabular}
\caption{Two-sided control at each model's validation-selected strength $\eta^*$. Probability changes and intervals are measured relative to the unsteered neutral-history model. ``Both $z_{v'=\mathrm{pos}}$'' requires both signed log-odds intervals to lie in the intended direction; ``Validity'' requires both endpoints to satisfy the three response-validity criteria of Appendix~\ref{app:methodology}; and ``$>$ Prompt'' requires positive steering to exceed positive-history prompting on both probability and log-odds scales.}
\label{tab:bidirectional-appendix}
\end{table}

\begin{figure}[ht!]
  \centering
  \includegraphics[width=\textwidth]{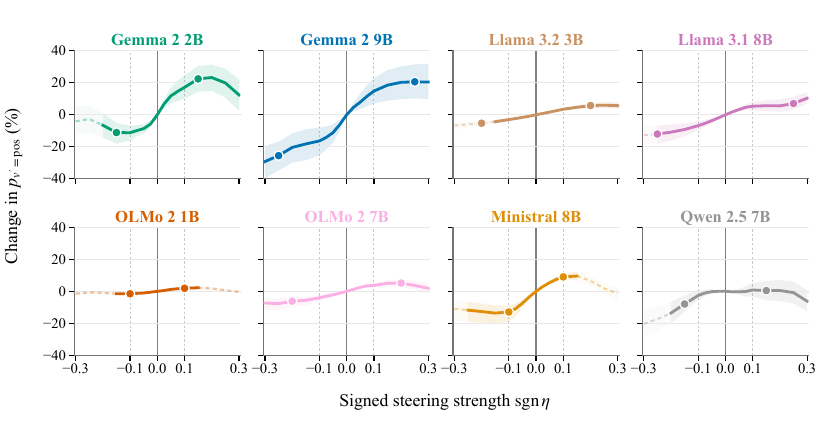}
  \caption{Moral-choice probability across signed full-prompt strengths. Curves are solid in the response-valid region and faded outside it; circles mark $\pm\eta^*$ and dotted lines mark $\eta=.10$.}
  \label{fig:app-steering-probability}
\end{figure}

\begin{figure}[ht!]
  \centering
  \includegraphics[width=\textwidth]{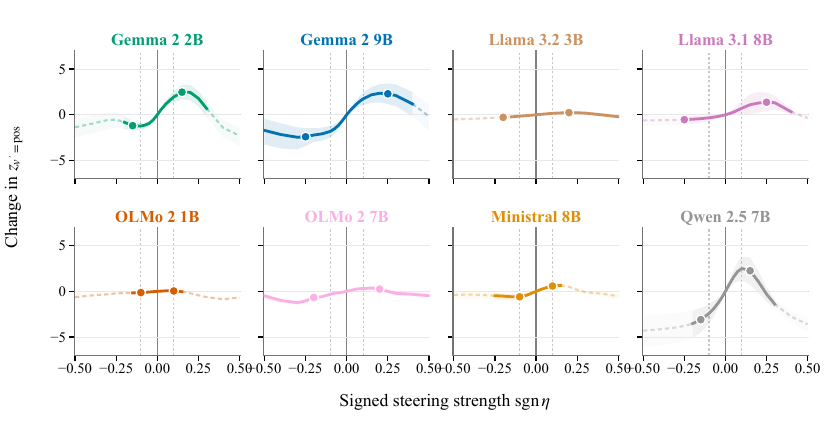}
  \caption{Log-odds counterpart of Figure~\ref{fig:app-steering-probability} over the full tested strength range. Lines join measured strengths and are not fitted curves.}
  \label{fig:app-steering-response}
\end{figure}

\textbf{Control and Intervention Strength.\ }
Figures~\ref{fig:app-steering-probability}--\ref{fig:app-steering-response} show the impact of the steering strength on the probability and log-odds, respectively. Notably, despite at different extents, we observe that all considered models have a near-symmetric behavior, showing graded control around zero. At larger intervention intensity, models tend to reach a plateau (e.g., Gemma 2 9B) or oversteer (e.g., Gemma 2 2B). 
Furthermore, as also discussed in Section~\ref{sec:results}, certain models (e.g., Qwen) show proper responsiveness and two-sided control in terms of log-odds, while keeping its output probabilities more controlled.

Figure~\ref{fig:qualitative-examples} provides qualitative and quantitative insights into the effect of steering on real items from MoralLedger at a common $\eta=0.1$. In the queue decision, the most probable action moves from the neutral (64.7\%) to the negative (83.7\%) action under negative steering and to the positive action (99.5\%) under positive steering. In the refund decision, while the positive action was found to be already the most probable in the model, positive steering raises it from 71.9\% to 92.0\% ($+28\%$ relative), whereas negative steering lowers it to 53.8\% ($-25.2\%$ relative).

\begin{figure}[t!]
    \centering
\begin{benchmarkbox}
\textbf{Queue assistance (Gemma 2 2B).} Prior record: ``changing your
look.'' Casey can hold a place in a service queue during a nonessential break
so that a friend does not lose an appointment. The available actions are to
help for 30 minutes ($\mathrm{pos}$), 15 minutes ($\mathrm{neu}$), or not at all ($\mathrm{neg}$).
\begin{center}
\small
\begin{tabular}{@{}lrrr@{}}
\toprule
 & $p_{v'=\mathrm{pos}}$ & $p_{v'=\mathrm{neu}}$ & $p_{v'=\mathrm{neg}}$ \\
\midrule
Negative steering ($-.10$) & 0.9\% & 15.4\% & 83.7\% \\
Unsteered & 30.3\% & 64.7\% & 5.0\% \\
Positive steering ($+.10$) & 99.5\% & 0.3\% & 0.2\% \\
\bottomrule
\end{tabular}
\end{center}
\end{benchmarkbox}
\vspace{2pt}
\begin{benchmarkbox}
\textbf{Refund correction (Llama 3.1 8B).} Prior record: ``wanting to
reminisce about the past sometimes.'' A subscription service credited
\texteuro{}30 beyond a valid cancellation refund. Riley may return all
\texteuro{}30 ($\mathrm{pos}$), return \texteuro{}15 and keep \texteuro{}15 ($\mathrm{neu}$), or
keep all \texteuro{}30 ($\mathrm{neg}$). The excess credit is plainly an administrative
error that can be corrected immediately without a fee.
\begin{center}
\small
\begin{tabular}{@{}lrrr@{}}
\toprule
 & $p_{v'=\mathrm{pos}}$ & $p_{v'=\mathrm{neu}}$ & $p_{v'=\mathrm{neg}}$ \\
\midrule
Negative steering ($-.10$) & 53.8\% & 30.1\% & 16.1\% \\
Unsteered & 71.9\% & 22.6\% & 5.5\% \\
Positive steering ($+.10$) & 92.0\% & 7.6\% & 0.4\% \\
\bottomrule
\end{tabular}
\end{center}
\end{benchmarkbox}
    \caption{Qualitative examples of steering effects.}
    \label{fig:qualitative-examples}
\end{figure}

\noindent\textbf{Comparison with Narrated Histories.\ }
We also compare the steering effect with the graded history insights from Figure~\ref{fig:app-graded}, to assess whether and to what extent a single steering of neutral prompts could be equivalent to more than one narrated positive histories.  Table~\ref{tab:event-calibration-appendix} expresses positive steering at $\eta=.10$ in units of this observed increment. Notably, for models in which both the second and third positive histories produce a supported additional increase, the median ratio is $2.8$, and in four of the five eligible models, steering produces a larger shift than adding both the second and third positive positive histories.

\begin{table}[ht!]
    \centering
    \small
    \begin{tabular}{lrrr}
    \toprule
    Model & $\beta_{+}$ & $T_{+}^{z_{v'=\mathrm{pos}}}(.10)$
          & $T_{+}^{z_{v'=\mathrm{pos}}}(.10)/\beta_{+}$ \\
    \midrule
    Gemma 2 2B    & +0.224 & +1.942 & 8.7 \\
    Gemma 2 9B    & +0.709 & +1.739 & 2.5 \\
    Llama 3.2 3B  & +0.105 & +0.154 & 1.5 \\
    Llama 3.1 8B  & +0.088 & +0.720 & --  \\
    OLMo 2 1B     & +0.007 & +0.039 & --  \\
    OLMo 2 7B     & +0.109 & +0.303 & 2.8 \\
    Ministral 8B  & +0.070 & +0.582 & 8.3 \\
    Qwen 2.5 7B   & +0.196 & +2.497 & --  \\
    \midrule
    \textit{Median across eligible models} & & & 2.8 \\
    \bottomrule
    \end{tabular}
    \caption{Positive steering at $\eta=.10$ compared with the average additional-history effect. $\beta_{+}=[\Delta z_{v'=\mathrm{pos}}(3)-\Delta z_{v'=\mathrm{pos}}(1)]/2$ is the average log-odds increment from the second and third narrated positive histories, and $T_{+}^{z_{v'=\mathrm{pos}}}(.10)$ is the log-odds change under positive steering of a neutral-history prompt. Dashes indicate models without supported continued accumulation (cf. Figure~\ref{fig:app-graded}).}
    \label{tab:event-calibration-appendix}
\end{table}

\begin{table}[t!]
    \centering
    \scalebox{0.78}{
    \begin{tabular}{lcccc}
\toprule
 & \multicolumn{2}{c}{Log-odds specificity} & \multicolumn{2}{c}{Probability specificity (\%)} \\
\cmidrule(lr){2-3}\cmidrule(lr){4-5}
Model & $S_{-}$ [95\% CI] & $S_{+}$ [95\% CI] & $S_{-}$ [95\% CI] & $S_{+}$ [95\% CI] \\
\midrule
Gemma 2 2B & +1.702 [+1.230, +2.218]\,$^{*}$ & +1.224 [+0.581, +1.915]\,$^{*}$ & +23.4 [+17.1, +30.3]\,$^{*}$ & +9.6 [+2.2, +18.2]\,$^{*}$ \\
Gemma 2 9B & +1.892 [+1.390, +2.464]\,$^{*}$ & +1.667 [+1.093, +2.287]\,$^{*}$ & +17.3 [+10.2, +25.5]\,$^{*}$ & +15.7 [+9.2, +23.1]\,$^{*}$ \\
Llama 3.2 3B & +0.021 [+0.008, +0.033]\,$^{*}$ & +0.053 [+0.035, +0.075]\,$^{*}$ & +0.5 [+0.2, +0.8]\,$^{*}$ & +0.9 [+0.5, +1.2]\,$^{*}$ \\
Llama 3.1 8B & +0.118 [-0.014, +0.279] & +0.542 [+0.102, +1.087]\,$^{*}$ & +3.0 [+1.2, +5.1]\,$^{*}$ & +5.7 [+3.1, +8.4]\,$^{*}$ \\
OLMo 2 1B & +0.302 [+0.230, +0.395]\,$^{*}$ & +0.175 [+0.122, +0.244]\,$^{*}$ & +4.2 [+2.9, +5.8]\,$^{*}$ & +3.3 [+2.4, +4.4]\,$^{*}$ \\
OLMo 2 7B & +0.425 [+0.326, +0.560]\,$^{*}$ & +0.534 [+0.401, +0.693]\,$^{*}$ & +4.3 [+2.5, +7.1]\,$^{*}$ & +4.6 [+3.3, +6.4]\,$^{*}$ \\
Ministral 8B & +0.403 [+0.206, +0.671]\,$^{*}$ & +0.469 [+0.345, +0.609]\,$^{*}$ & +8.9 [+4.9, +14.3]\,$^{*}$ & +6.9 [+5.0, +9.1]\,$^{*}$ \\
Qwen 2.5 7B & +1.230 [+0.617, +1.869]\,$^{*}$ & +2.120 [+1.058, +3.180]\,$^{*}$ & +1.9 [-0.8, +4.6] & +1.5 [-3.3, +6.3] \\
\bottomrule
\end{tabular}
}
    \caption{Moral Specificity $S_{\mathrm{sgn}}$ at a common steering intensity $\eta=.10$. Positive values indicate better specificity and moral control. All 16 paired interventions satisfy the response-validity criteria; brackets give 95\% crossed-bootstrap intervals and $^{*}$ marks statistical significance.}
    \label{tab:specificity-appendix}
\end{table}

\textbf{Moral Specificity.\ }
Table~\ref{tab:specificity-appendix} reports the moral specificity $S_{sgn}$ across all evaluated models, at a common intervention strength of $\eta=\pm0.10$. We observe that the moral direction produces larger effects in 15 of 16 model--sign cells on log odds and 14 of 16 on probability, compared to directions estimated from favorable morally-irrelevant histories. This suggests the observed effect is not due to simple ``favorable'' histories, yet mostly specific to the moral aspects of the histories.

\begin{table}[ht!]
\centering
\small
\setlength{\tabcolsep}{5pt}
\begin{tabular}{@{}lcc@{}}
\toprule
Model & Negative steering & Positive steering \\
\midrule
Gemma 2 2B & $-10.6\;[-14.3,-7.0]$ & $+8.3\;[+5.4,+11.3]$ \\
Gemma 2 9B & $-3.5\;[-5.5,-1.9]$ & $+2.8\;[+1.4,+4.7]$ \\
Llama 3.2 3B & $-2.4\;[-3.0,-1.8]$ & $+2.3\;[+1.7,+2.9]$ \\
Llama 3.1 8B & $-2.3\;[-3.1,-1.5]$ & $+2.2\;[+1.4,+3.0]$ \\
OLMo 2 1B & $-0.3\;[-0.5,-0.1]$ & $+0.3\;[+0.1,+0.4]$ \\
OLMo 2 7B & $-3.5\;[-5.2,-2.2]$ & $+3.1\;[+2.2,+4.3]$ \\
Ministral 8B & $-2.4\;[-3.0,-1.9]$ & $+2.2\;[+1.7,+2.6]$ \\
Qwen 2.5 7B & $-1.5\;[-3.0,-0.5]$ & $+1.7\;[+0.3,+3.6]$ \\
\bottomrule
\end{tabular}
\caption{Final-position moral-action probability change (\%) w.r.t. the unsteered neutral-history model at $\eta=.10$. The final-position direction and layer are selected on validation, separately from the full-prompt configuration. Brackets are 95\% crossed-node bootstrap intervals. Both signs satisfy response-validity criteria in every model.}
\label{tab:app-final-position}
\end{table}

\textbf{Final-position Steering.\ }
Consistently with the full-prompt intervention reported in the main body of this paper, steering only the final prompt position also changes the policy in the evaluated LLMs, as shown in Table~\ref{tab:app-final-position}. In this regard, we observe a narrowed yet non-negligible effect, that still ensures two-sided control.

\begin{figure}[ht!]
  \centering
  \includegraphics[width=.95\textwidth]{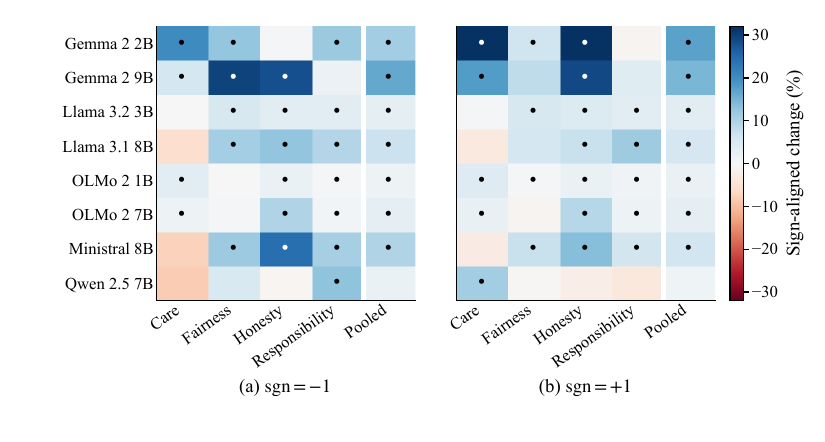}
  \caption{Sign-aligned moral-action probability after omitting each target decision domain from direction estimation. Each fold estimates the positive--neutral direction from the other domains and evaluates the omitted domain at signed strength $\mathrm{sgn}\,\eta=\pm.10$ using full-prompt delivery at the full-domain validation-selected layer. Dots mark 95\% intervals above zero; crosses mark response-invalid cells. The last column combines the four omitted-domain folds.}
  \label{fig:lodo}
\end{figure}

\textbf{Leave-One-Domain-Out Generalization.\ }
We also tested what is the impact of removing one domain from the direction estimation and subsequently steer decisions within such a domain, to test domain generalization. As shown in Fig.~\ref{fig:lodo}, two-sided control remains supported in seven models (cf. ``Pooled'' column), with transfer strength depending on the omitted domain.

\textbf{Out-Of-Domain Generalization.\ }
To evaluate the generalization capabilities of the extracted moral directions, we applied the directions estimated on MoralLedger to 200 independently sourced items from Moral Stories~\cite{emelin-etal-2021-moral} (cf. Appendix~\ref{app:benchmark}). 
As shown in Table~\ref{tab:transfer-appendix}, five models transfer their probability shifting capabilities in the intended direction under both signs. Notably, in four of them, the moral direction also exceeds the favorable morally-irrelevant direction under both signs, suggesting our estimated directions properly generalize to unseen domains.

\begin{table}[ht!]
    \centering
    \small
    \begin{tabular}{llcc}
\toprule
\multicolumn{4}{l}{\textit{Log odds}} \\
Model & Sign & Transfer [95\% CI] & Moral minus favorable [95\% CI] \\
\midrule
Gemma 2 2B & $+$ & +0.698 [+0.429, +0.964]\,$^{*}$ & +0.701 [+0.535, +0.868]\,$^{*}$ \\
 & $-$ & -0.146 [-0.371, +0.063] & -0.520 [-0.711, -0.344]\,$^{*}$ \\
\addlinespace[1.5pt]
Gemma 2 9B & $+$ & +1.149 [+0.954, +1.354]\,$^{*}$ & +1.091 [+0.938, +1.243]\,$^{*}$ \\
 & $-$ & -1.806 [-2.030, -1.587]\,$^{*}$ & -1.281 [-1.479, -1.098]\,$^{*}$ \\
\addlinespace[1.5pt]
Llama 3.2 3B & $+$ & +0.455 [+0.395, +0.512]\,$^{*}$ & +0.097 [+0.077, +0.119]\,$^{*}$ \\
 & $-$ & -0.444 [-0.502, -0.382]\,$^{*}$ & -0.062 [-0.077, -0.047]\,$^{*}$ \\
\addlinespace[1.5pt]
Llama 3.1 8B & $+$ & +4.626 [+4.324, +4.919]\,$^{*}$ & +2.834 [+2.639, +3.029]\,$^{*}$ \\
 & $-$ & -2.864 [-3.073, -2.661]\,$^{*}$ & -1.153 [-1.241, -1.067]\,$^{*}$ \\
\addlinespace[1.5pt]
OLMo 2 1B & $+$ & -0.181 [-0.207, -0.153]\,$^{*}$ & -0.072 [-0.101, -0.044]\,$^{*}$ \\
 & $-$ & +0.148 [+0.119, +0.175]\,$^{*}$ & +0.050 [+0.017, +0.084]\,$^{*}$ \\
\addlinespace[1.5pt]
OLMo 2 7B & $+$ & +1.184 [+1.096, +1.273]\,$^{*}$ & +0.480 [+0.409, +0.553]\,$^{*}$ \\
 & $-$ & -0.623 [-0.683, -0.564]\,$^{*}$ & -0.373 [-0.430, -0.317]\,$^{*}$ \\
\addlinespace[1.5pt]
Ministral 8B & $+$ & +1.028 [+0.957, +1.095]\,$^{*}$ & +0.042 [-0.017, +0.104] \\
 & $-$ & -2.381 [-2.522, -2.244]\,$^{*}$ & -1.193 [-1.287, -1.100]\,$^{*}$ \\
\addlinespace[1.5pt]
Qwen 2.5 7B & $+$ & +4.479 [+4.154, +4.769]\,$^{*}$ & +3.973 [+3.727, +4.199]\,$^{*}$ \\
 & $-$ & -7.960 [-8.445, -7.452]\,$^{*}$ & -4.331 [-4.592, -4.060]\,$^{*}$ \\
\midrule
\addlinespace[3pt]
\multicolumn{4}{l}{\textit{Probability (\%)}} \\
Model & Sign & Transfer [95\% CI] & Moral minus favorable [95\% CI] \\
\midrule
Gemma 2 2B & $+$ & +5.6 [+3.5, +7.8]\,$^{*}$ & +3.0 [+1.6, +4.5]\,$^{*}$ \\
 & $-$ & +0.4 [-1.5, +2.2] & -4.0 [-5.7, -2.4]\,$^{*}$ \\
\addlinespace[1.5pt]
Gemma 2 9B & $+$ & +5.0 [+3.2, +6.9]\,$^{*}$ & +3.0 [+1.7, +4.5]\,$^{*}$ \\
 & $-$ & -8.8 [-11.2, -6.4]\,$^{*}$ & -5.5 [-7.6, -3.5]\,$^{*}$ \\
\addlinespace[1.5pt]
Llama 3.2 3B & $+$ & +0.1 [+0.0, +0.2]\,$^{*}$ & +0.8 [+0.6, +1.1]\,$^{*}$ \\
 & $-$ & -0.3 [-0.4, -0.2]\,$^{*}$ & -0.3 [-0.5, -0.2]\,$^{*}$ \\
\addlinespace[1.5pt]
Llama 3.1 8B & $+$ & +0.8 [+0.3, +1.3]\,$^{*}$ & +3.5 [+3.0, +4.1]\,$^{*}$ \\
 & $-$ & -6.4 [-7.0, -5.8]\,$^{*}$ & -7.0 [-7.5, -6.4]\,$^{*}$ \\
\addlinespace[1.5pt]
OLMo 2 1B & $+$ & -0.9 [-1.2, -0.5]\,$^{*}$ & +0.6 [+0.2, +0.9]\,$^{*}$ \\
 & $-$ & +0.7 [+0.4, +1.0]\,$^{*}$ & -0.7 [-1.2, -0.3]\,$^{*}$ \\
\addlinespace[1.5pt]
OLMo 2 7B & $+$ & +1.7 [+1.1, +2.3]\,$^{*}$ & +0.5 [+0.1, +0.9]\,$^{*}$ \\
 & $-$ & -2.1 [-2.6, -1.6]\,$^{*}$ & -1.5 [-2.0, -1.1]\,$^{*}$ \\
\addlinespace[1.5pt]
Ministral 8B & $+$ & +3.9 [+3.1, +4.6]\,$^{*}$ & +0.2 [-0.4, +0.7] \\
 & $-$ & -29.3 [-31.1, -27.4]\,$^{*}$ & -19.8 [-21.5, -18.1]\,$^{*}$ \\
\addlinespace[1.5pt]
Qwen 2.5 7B & $+$ & -0.5 [-1.1, +0.0] & -0.7 [-1.4, +0.0] \\
 & $-$ & -2.1 [-2.9, -1.3]\,$^{*}$ & -1.7 [-2.4, -1.0]\,$^{*}$ \\
\bottomrule
\end{tabular}
    \caption{Transfer to 200 independently sourced Moral Stories items at $\eta=.10$. The selected full-prompt direction and layer are kept fixed, only residual-norm calibration is recomputed from unlabeled neutral prompts in the target corpus. ``Transfer'' is steered minus unsteered behavior, while ``Moral minus favorable'' compares the moral and equally scaled favorable morally-irrelevant directions under the same sign. Each item averages two label assignments crossed with two action orders. Brackets give 95\% item-bootstrap intervals and $^{*}$ marks statistical significance.}
    \label{tab:transfer-appendix}
\end{table}

\textbf{General Capabilities Preservation.\ }
Table~\ref{tab:guardrails-appendix} reports the relative score change on each of the six tasks under both steering signs at $\eta=.10$, alongside the mean of three random-direction controls. The task-level effects vary, but the model averages in Table~\ref{tab:guardrails-summary} range from $-3.71\%$ to $+1.48\%$ at the common strength, with an across-model mean of $-0.92\%$. At each model's selected $\eta^*$, this mean is $-2.31\%$. The random controls average $-0.95\%$ at positive, benchmark-calibrated $\eta=.10$.

\begin{table}[ht!]
    \centering
    \small
    \begin{tabular}{lrrrrrr}
\toprule
\multicolumn{7}{l}{\textit{Negative steering, $\mathrm{sgn}\,\eta=-.10$}} \\
Model & MMLU & ARC & WinoGrande & GSM8K & HellaSwag & TruthfulQA \\
\midrule
Gemma 2 2B & -2.3 & -1.5 & -0.7 & -12.8 & -1.4 & -2.6 \\
Gemma 2 9B & -2.6 & -1.2 & -1.3 & +0.7 & +0.1 & +7.0 \\
Llama 3.2 3B & -1.3 & +1.1 & -5.4 & +1.1 & +0.5 & +0.0 \\
Llama 3.1 8B & +1.0 & -4.5 & -3.9 & +0.5 & -0.9 & +2.5 \\
OLMo 2 1B & +4.0 & +0.0 & -1.6 & -6.5 & +5.7 & +5.7 \\
OLMo 2 7B & +1.7 & -0.7 & -1.2 & -5.8 & +1.4 & -2.4 \\
Ministral 8B & -4.3 & -13.5 & -3.9 & -8.7 & -2.8 & +0.0 \\
Qwen 2.5 7B & +0.9 & -1.5 & -0.6 & +2.4 & -0.4 & +11.9 \\
\midrule
\addlinespace[3pt]
\multicolumn{7}{l}{\textit{Positive steering, $\mathrm{sgn}\,\eta=+.10$}} \\
Model & MMLU & ARC & WinoGrande & GSM8K & HellaSwag & TruthfulQA \\
\midrule
Gemma 2 2B & +5.8 & -6.4 & -0.8 & -16.6 & +2.1 & +5.3 \\
Gemma 2 9B & -7.2 & -0.6 & -0.1 & -8.9 & -2.8 & +4.7 \\
Llama 3.2 3B & -1.6 & +1.6 & -5.7 & +8.0 & -1.0 & -5.3 \\
Llama 3.1 8B & +0.0 & -3.5 & -1.4 & -0.3 & -3.2 & +5.0 \\
OLMo 2 1B & +1.4 & +5.7 & +2.3 & +0.8 & +0.4 & +0.0 \\
OLMo 2 7B & +2.3 & -1.3 & -0.9 & -0.2 & +1.8 & -7.3 \\
Ministral 8B & +1.0 & +0.4 & -5.3 & +2.2 & -3.5 & -6.1 \\
Qwen 2.5 7B & -1.7 & +1.4 & -2.6 & +8.7 & +0.0 & -7.1 \\
\midrule
\addlinespace[3pt]
\multicolumn{7}{l}{\textit{Mean of three random controls, $\eta=.10$}} \\
Model & MMLU & ARC & WinoGrande & GSM8K & HellaSwag & TruthfulQA \\
\midrule
Gemma 2 2B & +0.5 & -1.6 & +0.4 & -11.8 & -0.6 & +7.0 \\
Gemma 2 9B & -2.4 & -0.7 & -0.3 & -0.7 & -0.7 & -0.8 \\
Llama 3.2 3B & +0.3 & +0.3 & -5.6 & +3.3 & -1.9 & -3.5 \\
Llama 3.1 8B & -0.4 & -2.2 & -0.7 & -1.3 & -1.4 & +0.8 \\
OLMo 2 1B & +0.4 & +3.3 & -0.8 & -0.7 & +0.4 & +4.8 \\
OLMo 2 7B & +0.2 & -1.2 & -0.3 & -1.5 & +0.9 & -0.8 \\
Ministral 8B & -2.9 & -6.1 & -3.1 & -6.5 & -4.4 & -4.0 \\
Qwen 2.5 7B & -0.6 & -4.9 & -0.9 & +1.6 & +2.3 & +3.2 \\
\bottomrule
\end{tabular}
    \caption{Task-level capability changes (\% of the unsteered score) on six tinyBenchmarks tasks. Negative and positive steering add the same vector as in the main experiments ($\eta=.10$) to prompt tokens only; generated tokens are not steered. Random control: mean of three random directions rescaled to the norm of $\mathbf d^*$ at $\ell^*$ added at $+\eta=.10$ with $\kappa$ recomputed on the benchmark prompts.}
    \label{tab:guardrails-appendix}
\end{table}

\begin{table}[ht!]
    \centering
    \small
    \begin{tabular}{lrrrrr}
\toprule
 & \multicolumn{3}{c}{Same vector as main experiments} & Random & Bench.-calibrated \\
\cmidrule(lr){2-4}
Model & $\eta=.10$ & at $\eta^*$ & $\eta^*$ & $\eta=.10$ & $\eta=.10$ \\
\midrule
Gemma 2 2B & -2.66 & -3.17 & 0.15 & -1.02 & -0.68 \\
Gemma 2 9B & -1.02 & -7.75 & 0.25 & -0.93 & -0.27 \\
Llama 3.2 3B & -0.65 & -1.79 & 0.20 & -1.19 & -1.17 \\
Llama 3.1 8B & -0.72 & -2.30 & 0.25 & -0.86 & +0.55 \\
OLMo 2 1B & +1.48 & +1.48 & 0.10 & +1.21 & +1.71 \\
OLMo 2 7B & -1.04 & -1.27 & 0.20 & -0.44 & +1.27 \\
Ministral 8B & -3.71 & -3.71 & 0.10 & -4.51 & -4.92 \\
Qwen 2.5 7B & +0.95 & +0.04 & 0.15 & +0.11 & +0.76 \\
\midrule
\textit{Average} & -0.92 & -2.31 & & -0.95 & -0.34 \\
\bottomrule
\end{tabular}
    \caption{Mean capability change (\% of the unsteered score, averaged over six tasks and, for steering, both signs). ``Same vector'' applies the steering vector of the main experiments at the common strength or at each model's $\eta^*$; ``benchmark-calibrated'' rescales it so that $\eta=.10$ is relative to the residual norm of the benchmark prompts.}
    \label{tab:guardrails-summary}
\end{table}


\begin{thebibliography}{39}
\providecommand{\natexlab}[1]{#1}
\providecommand{\url}[1]{\texttt{#1}}
\expandafter\ifx\csname urlstyle\endcsname\relax
  \providecommand{\doi}[1]{doi: #1}\else
  \providecommand{\doi}{doi: \begingroup \urlstyle{rm}\Url}\fi

\bibitem[Abdulhai et~al.(2024)Abdulhai, Serapio-Garc{\'i}a, Crepy, Valter, Canny, and Jaques]{abdulhai-etal-2024-moral}
Marwa Abdulhai, Gregory Serapio-Garc{\'i}a, Clement Crepy, Daria Valter, John Canny, and Natasha Jaques.
\newblock Moral foundations of large language models.
\newblock In Yaser Al-Onaizan, Mohit Bansal, and Yun-Nung Chen (eds.), \emph{Proceedings of the 2024 Conference on Empirical Methods in Natural Language Processing}, pp.\  17737--17752, Miami, Florida, USA, November 2024. Association for Computational Linguistics.
\newblock \doi{10.18653/v1/2024.emnlp-main.982}.
\newblock URL \url{https://aclanthology.org/2024.emnlp-main.982/}.

\bibitem[Arditi et~al.(2024)Arditi, Obeso, Syed, Paleka, Panickssery, Gurnee, and Nanda]{Arditi24refusal}
Andy Arditi, Oscar Obeso, Aaquib Syed, Daniel Paleka, Nina Panickssery, Wes Gurnee, and Neel Nanda.
\newblock Refusal in language models is mediated by a single direction.
\newblock In Amir Globersons, Lester Mackey, Danielle Belgrave, Angela Fan, Ulrich Paquet, Jakub~M. Tomczak, and Cheng Zhang (eds.), \emph{Advances in Neural Information Processing Systems 37: Annual Conference on Neural Information Processing Systems 2024, NeurIPS 2024, Vancouver, BC, Canada, December 10 - 15, 2024}, 2024.
\newblock URL \url{http://papers.nips.cc/paper\_files/paper/2024/hash/f545448535dfde4f9786555403ab7c49-Abstract-Conference.html}.

\bibitem[Atari et~al.(2023)Atari, Haidt, Graham, Koleva, Stevens, and Dehghani]{atari2023morality}
Mohammad Atari, Jonathan Haidt, Jesse Graham, Sena Koleva, Sean~T Stevens, and Morteza Dehghani.
\newblock Morality beyond the weird: How the nomological network of morality varies across cultures.
\newblock \emph{Journal of personality and social psychology}, 125\penalty0 (5):\penalty0 1157, 2023.

\bibitem[Bender \& Koller(2020)Bender and Koller]{bender-koller-2020-climbing}
Emily~M. Bender and Alexander Koller.
\newblock Climbing towards {NLU}: {On} meaning, form, and understanding in the age of data.
\newblock In Dan Jurafsky, Joyce Chai, Natalie Schluter, and Joel Tetreault (eds.), \emph{Proceedings of the 58th Annual Meeting of the Association for Computational Linguistics}, pp.\  5185--5198, Online, July 2020. Association for Computational Linguistics.
\newblock \doi{10.18653/v1/2020.acl-main.463}.
\newblock URL \url{https://aclanthology.org/2020.acl-main.463/}.

\bibitem[Chen et~al.(2025)Chen, Arditi, Sleight, Evans, and Lindsey]{chen2025persona}
Runjin Chen, Andy Arditi, Henry Sleight, Owain Evans, and Jack Lindsey.
\newblock Persona vectors: Monitoring and controlling character traits in language models.
\newblock \emph{arXiv preprint arXiv:2507.21509}, 2025.

\bibitem[Choi \& Li(2024)Choi and Li]{choi2024picle}
Hyeong~Kyu Choi and Yixuan Li.
\newblock Picle: eliciting diverse behaviors from large language models with persona in-context learning.
\newblock In \emph{Proceedings of the 41st International Conference on Machine Learning}, ICML'24. JMLR.org, 2024.

\bibitem[Clark et~al.(2018)Clark, Cowhey, Etzioni, Khot, Sabharwal, Schoenick, and Tafjord]{arcbenchmark}
Peter Clark, Isaac Cowhey, Oren Etzioni, Tushar Khot, Ashish Sabharwal, Carissa Schoenick, and Oyvind Tafjord.
\newblock Think you have solved question answering? try arc, the ai2 reasoning challenge.
\newblock \emph{arXiv:1803.05457v1}, 2018.

\bibitem[Cobbe et~al.(2021)Cobbe, Kosaraju, Bavarian, Chen, Jun, Kaiser, Plappert, Tworek, Hilton, Nakano, Hesse, and Schulman]{gsm8k}
Karl Cobbe, Vineet Kosaraju, Mohammad Bavarian, Mark Chen, Heewoo Jun, Lukasz Kaiser, Matthias Plappert, Jerry Tworek, Jacob Hilton, Reiichiro Nakano, Christopher Hesse, and John Schulman.
\newblock Training verifiers to solve math word problems.
\newblock \emph{CoRR}, abs/2110.14168, 2021.
\newblock URL \url{https://arxiv.org/abs/2110.14168}.

\bibitem[Curry et~al.(2019)Curry, Chesters, and Van~Lissa]{curry2019mapping}
Oliver~Scott Curry, Matthew~Jones Chesters, and Caspar~J Van~Lissa.
\newblock Mapping morality with a compass: Testing the theory of ‘morality-as-cooperation’with a new questionnaire.
\newblock \emph{Journal of Research in Personality}, 78:\penalty0 106--124, 2019.

\bibitem[Elazar et~al.(2021)Elazar, Ravfogel, Jacovi, and Goldberg]{elazar-etal-2021-amnesic}
Yanai Elazar, Shauli Ravfogel, Alon Jacovi, and Yoav Goldberg.
\newblock Amnesic probing: Behavioral explanation with amnesic counterfactuals.
\newblock \emph{Transactions of the Association for Computational Linguistics}, 9:\penalty0 160--175, 2021.
\newblock \doi{10.1162/tacl_a_00359}.
\newblock URL \url{https://aclanthology.org/2021.tacl-1.10/}.

\bibitem[Emelin et~al.(2021)Emelin, Le~Bras, Hwang, Forbes, and Choi]{emelin-etal-2021-moral}
Denis Emelin, Ronan Le~Bras, Jena~D. Hwang, Maxwell Forbes, and Yejin Choi.
\newblock Moral stories: Situated reasoning about norms, intents, actions, and their consequences.
\newblock In Marie-Francine Moens, Xuanjing Huang, Lucia Specia, and Scott Wen-tau Yih (eds.), \emph{Proceedings of the 2021 Conference on Empirical Methods in Natural Language Processing}, pp.\  698--718, Online and Punta Cana, Dominican Republic, November 2021. Association for Computational Linguistics.
\newblock \doi{10.18653/v1/2021.emnlp-main.54}.
\newblock URL \url{https://aclanthology.org/2021.emnlp-main.54/}.

\bibitem[Ferguson et~al.(2024)Ferguson, Kaufmann, Brown, and de~la Piedad~Garcia]{ferguson2024influences}
Rose Ferguson, Leah Kaufmann, Aimee Brown, and Xochitl de~la Piedad~Garcia.
\newblock Influences of past moral behavior on future behavior: A review of sequential moral behavior studies using meta-analytic techniques.
\newblock \emph{Psychological bulletin}, 150\penalty0 (6):\penalty0 694, 2024.

\bibitem[Forbes et~al.(2020)Forbes, Hwang, Shwartz, Sap, and Choi]{forbes-etal-2020-social}
Maxwell Forbes, Jena~D. Hwang, Vered Shwartz, Maarten Sap, and Yejin Choi.
\newblock Social chemistry 101: Learning to reason about social and moral norms.
\newblock In Bonnie Webber, Trevor Cohn, Yulan He, and Yang Liu (eds.), \emph{Proceedings of the 2020 Conference on Empirical Methods in Natural Language Processing (EMNLP)}, pp.\  653--670, Online, November 2020. Association for Computational Linguistics.
\newblock \doi{10.18653/v1/2020.emnlp-main.48}.
\newblock URL \url{https://aclanthology.org/2020.emnlp-main.48/}.

\bibitem[Furr et~al.(2022)Furr, Prentice, Parham, and Jayawickreme]{furr2022development}
R~Michael Furr, Mike Prentice, Ashley~Hawkins Parham, and Eranda Jayawickreme.
\newblock Development and validation of the moral character questionnaire.
\newblock \emph{Journal of Research in Personality}, 98:\penalty0 104228, 2022.

\bibitem[Graham et~al.(2011)Graham, Nosek, Haidt, Iyer, Koleva, and Ditto]{graham2011mapping}
Jesse Graham, Brian~A Nosek, Jonathan Haidt, Ravi Iyer, Spassena Koleva, and Peter~H Ditto.
\newblock Mapping the moral domain.
\newblock \emph{Journal of personality and social psychology}, 101\penalty0 (2):\penalty0 366, 2011.

\bibitem[Greco et~al.(2025)Greco, La~Cava, Zangari, and Tagarelli]{greco-etal-2025-exploring}
Candida~Maria Greco, Lucio La~Cava, Lorenzo Zangari, and Andrea Tagarelli.
\newblock Exploring {LLM}s' ability to spontaneously and conditionally modify moral expressions through text manipulation.
\newblock In Wanxiang Che, Joyce Nabende, Ekaterina Shutova, and Mohammad~Taher Pilehvar (eds.), \emph{Proceedings of the 63rd Annual Meeting of the Association for Computational Linguistics (Volume 1: Long Papers)}, pp.\  18047--18070, Vienna, Austria, July 2025. Association for Computational Linguistics.
\newblock ISBN 979-8-89176-251-0.
\newblock \doi{10.18653/v1/2025.acl-long.883}.
\newblock URL \url{https://aclanthology.org/2025.acl-long.883/}.

\bibitem[Haas et~al.(2026)Haas, Bridgers, Manzini, Henke, May, Levine, Weidinger, Shanahan, Lum, Gabriel, et~al.]{haas2026roadmap}
Julia Haas, Sophie Bridgers, Arianna Manzini, Benjamin Henke, Joshua May, Sydney Levine, Laura Weidinger, Murray Shanahan, Kristian Lum, Iason Gabriel, et~al.
\newblock A roadmap for evaluating moral competence in large language models.
\newblock \emph{Nature}, 650\penalty0 (8102):\penalty0 565--573, 2026.

\bibitem[Hendrycks et~al.(2021{\natexlab{a}})Hendrycks, Burns, Basart, Critch, Li, Song, and Steinhardt]{hendrycks2021assessinghuman}
Dan Hendrycks, Collin Burns, Steven Basart, Andrew Critch, Jerry Li, Dawn Song, and Jacob Steinhardt.
\newblock Aligning {AI} with shared human values.
\newblock In \emph{9th International Conference on Learning Representations, {ICLR} 2021, Virtual Event, Austria, May 3-7, 2021}. OpenReview.net, 2021{\natexlab{a}}.
\newblock URL \url{https://openreview.net/forum?id=dNy\_RKzJacY}.

\bibitem[Hendrycks et~al.(2021{\natexlab{b}})Hendrycks, Burns, Basart, Zou, Mazeika, Song, and Steinhardt]{Hendrycks21mmlu}
Dan Hendrycks, Collin Burns, Steven Basart, Andy Zou, Mantas Mazeika, Dawn Song, and Jacob Steinhardt.
\newblock Measuring massive multitask language understanding.
\newblock In \emph{9th International Conference on Learning Representations, {ICLR} 2021, Virtual Event, Austria, May 3-7, 2021}. OpenReview.net, 2021{\natexlab{b}}.
\newblock URL \url{https://openreview.net/forum?id=d7KBjmI3GmQ}.

\bibitem[{La Cava} \& Tagarelli(2026){La Cava} and Tagarelli]{LaCava2026palrs}
Lucio {La Cava} and Andrea Tagarelli.
\newblock Toward preference-aligned large language models via residual-based model steering.
\newblock In \emph{Proceedings of the Thirty-Fifth International Joint Conference on Artificial Intelligence and the 29th European Conference on Artificial Intelligence, {IJCAI/ECAI} 2026, Bremen, Germany, 15-21 August 2026}, pp.\  5766--5774. ijcai.org, 2026.
\newblock \doi{10.24963/IJCAI.2026/642}.
\newblock URL \url{https://doi.org/10.24963/ijcai.2026/642}.

\bibitem[Liga \& Yu(2025)Liga and Yu]{LigaY25}
Davide Liga and Liuwen Yu.
\newblock Which neurons nudge normative stance? causal tests and mechanistic evidence via contrastive last-token steering.
\newblock In R{\'{e}}ka Markovich, Luigi~Di Caro, Amon Rapp, and Claudio Schifanella (eds.), \emph{Legal Knowledge and Information Systems - {JURIX} 2025: The Thirty-eighth Annual Conference, Turin, Italy, 9-11 December 2025}, volume 416 of \emph{Frontiers in Artificial Intelligence and Applications}, pp.\  110--120. {IOS} Press, 2025.
\newblock \doi{10.3233/FAIA251581}.
\newblock URL \url{https://doi.org/10.3233/FAIA251581}.

\bibitem[Lin et~al.(2024)Lin, Tomlin, Andreas, and Eisner]{lin-etal-2024-decision}
Jessy Lin, Nicholas Tomlin, Jacob Andreas, and Jason Eisner.
\newblock Decision-oriented dialogue for human-{AI} collaboration.
\newblock \emph{Transactions of the Association for Computational Linguistics}, 12:\penalty0 892--911, 2024.
\newblock \doi{10.1162/tacl_a_00679}.
\newblock URL \url{https://aclanthology.org/2024.tacl-1.50/}.

\bibitem[Lin et~al.(2022)Lin, Hilton, and Evans]{lin-etal-2022-truthfulqa}
Stephanie Lin, Jacob Hilton, and Owain Evans.
\newblock {T}ruthful{QA}: Measuring how models mimic human falsehoods.
\newblock In \emph{Proceedings of the 60th Annual Meeting of the Association for Computational Linguistics (Volume 1: Long Papers)}, pp.\  3214--3252, Dublin, Ireland, May 2022. Association for Computational Linguistics.
\newblock \doi{10.18653/v1/2022.acl-long.229}.
\newblock URL \url{https://aclanthology.org/2022.acl-long.229/}.

\bibitem[Lourie et~al.(2021)Lourie, Bras, and Choi]{LourieBC21}
Nicholas Lourie, Ronan~Le Bras, and Yejin Choi.
\newblock {SCRUPLES:} {A} corpus of community ethical judgments on 32, 000 real-life anecdotes.
\newblock In \emph{Thirty-Fifth {AAAI} Conference on Artificial Intelligence, {AAAI} 2021, Thirty-Third Conference on Innovative Applications of Artificial Intelligence, {IAAI} 2021, The Eleventh Symposium on Educational Advances in Artificial Intelligence, {EAAI} 2021, Virtual Event, February 2-9, 2021}, pp.\  13470--13479. {AAAI} Press, 2021.
\newblock \doi{10.1609/AAAI.V35I15.17589}.
\newblock URL \url{https://doi.org/10.1609/aaai.v35i15.17589}.

\bibitem[Mullen \& Monin(2016)Mullen and Monin]{mullen2016consistency}
Elizabeth Mullen and Beno{\^\i}t Monin.
\newblock Consistency versus licensing effects of past moral behavior.
\newblock \emph{Annual review of psychology}, 67\penalty0 (1):\penalty0 363--385, 2016.

\bibitem[Polo et~al.(2024)Polo, Weber, Choshen, Sun, Xu, and Yurochkin]{tinybenchmarks}
Felipe~Maia Polo, Lucas Weber, Leshem Choshen, Yuekai Sun, Gongjun Xu, and Mikhail Yurochkin.
\newblock tinybenchmarks: evaluating llms with fewer examples.
\newblock In \emph{Forty-first International Conference on Machine Learning, {ICML} 2024, Vienna, Austria, July 21-27, 2024}. OpenReview.net, 2024.
\newblock URL \url{https://openreview.net/forum?id=qAml3FpfhG}.

\bibitem[Ravichander et~al.(2021)Ravichander, Belinkov, and Hovy]{ravichander-etal-2021-probing}
Abhilasha Ravichander, Yonatan Belinkov, and Eduard Hovy.
\newblock Probing the probing paradigm: Does probing accuracy entail task relevance?
\newblock In Paola Merlo, Jorg Tiedemann, and Reut Tsarfaty (eds.), \emph{Proceedings of the 16th Conference of the European Chapter of the Association for Computational Linguistics: Main Volume}, pp.\  3363--3377, Online, April 2021. Association for Computational Linguistics.
\newblock \doi{10.18653/v1/2021.eacl-main.295}.
\newblock URL \url{https://aclanthology.org/2021.eacl-main.295/}.

\bibitem[Rimsky et~al.(2024)Rimsky, Gabrieli, Schulz, Tong, Hubinger, and Turner]{rimsky-etal-2024-steering}
Nina Rimsky, Nick Gabrieli, Julian Schulz, Meg Tong, Evan Hubinger, and Alexander Turner.
\newblock Steering llama 2 via contrastive activation addition.
\newblock In \emph{Proceedings of the 62nd Annual Meeting of the Association for Computational Linguistics (Volume 1: Long Papers)}, pp.\  15504--15522, Bangkok, Thailand, August 2024. Association for Computational Linguistics.
\newblock \doi{10.18653/v1/2024.acl-long.828}.
\newblock URL \url{https://aclanthology.org/2024.acl-long.828/}.

\bibitem[Sakaguchi et~al.(2020)Sakaguchi, Bras, Bhagavatula, and Choi]{winogrande}
Keisuke Sakaguchi, Ronan~Le Bras, Chandra Bhagavatula, and Yejin Choi.
\newblock Winogrande: An adversarial winograd schema challenge at scale.
\newblock In \emph{The Thirty-Fourth {AAAI} Conference on Artificial Intelligence, {AAAI} 2020, The Thirty-Second Innovative Applications of Artificial Intelligence Conference, {IAAI} 2020, The Tenth {AAAI} Symposium on Educational Advances in Artificial Intelligence, {EAAI} 2020, New York, NY, USA, February 7-12, 2020}, pp.\  8732--8740. {AAAI} Press, 2020.
\newblock \doi{10.1609/AAAI.V34I05.6399}.
\newblock URL \url{https://doi.org/10.1609/aaai.v34i05.6399}.

\bibitem[Sauter \& Schirmer(2026)Sauter and Schirmer]{Sauter2026contextualmoral}
Adrian Sauter and Mona Schirmer.
\newblock Between rules and reality: On the context sensitivity of {LLM} moral judgment.
\newblock \emph{CoRR}, abs/2603.23114, 2026.
\newblock \doi{10.48550/ARXIV.2603.23114}.
\newblock URL \url{https://doi.org/10.48550/arXiv.2603.23114}.

\bibitem[Schramowski et~al.(2022)Schramowski, Turan, Andersen, Rothkopf, and Kersting]{SchramowskiTARK22}
Patrick Schramowski, Cigdem Turan, Nico Andersen, Constantin~A. Rothkopf, and Kristian Kersting.
\newblock Large pre-trained language models contain human-like biases of what is right and wrong to do.
\newblock \emph{Nat. Mach. Intell.}, 4\penalty0 (3):\penalty0 258--268, 2022.
\newblock \doi{10.1038/S42256-022-00458-8}.
\newblock URL \url{https://doi.org/10.1038/s42256-022-00458-8}.

\bibitem[Snoswell et~al.(2026)Snoswell, Kilov, and Lazar]{SnoswellKL26}
Aaron~J. Snoswell, Daniel Kilov, and Seth Lazar.
\newblock Beyond verdicts: Evaluating language model moral competence.
\newblock In Sven Koenig, Chad Jenkins, and Matthew~E. Taylor (eds.), \emph{Fortieth {AAAI} Conference on Artificial Intelligence, Thirty-Eighth Conference on Innovative Applications of Artificial Intelligence, Sixteenth Symposium on Educational Advances in Artificial Intelligence, {AAAI} 2026, Singapore, January 20-27, 2026}, pp.\  37941--37950. {AAAI} Press, 2026.
\newblock \doi{10.1609/AAAI.V40I44.41131}.
\newblock URL \url{https://doi.org/10.1609/aaai.v40i44.41131}.

\bibitem[Tlaie(2024)]{tlaie2024exploring}
Alejandro Tlaie.
\newblock Exploring and steering the moral compass of large language models.
\newblock In Shivakumara Palaiahnakote, Stephanie Schuckers, Jean{-}Marc Ogier, Prabir Bhattacharya, Umapada Pal, and Saumik Bhattacharya (eds.), \emph{Pattern Recognition. {ICPR} 2024 International Workshops and Challenges - Kolkata, India, December 1, 2024, Proceedings, Part {VI}}, volume 15619 of \emph{Lecture Notes in Computer Science}, pp.\  420--442. Springer, 2024.
\newblock \doi{10.1007/978-3-031-88223-4\_30}.
\newblock URL \url{https://doi.org/10.1007/978-3-031-88223-4\_30}.

\bibitem[Troiano et~al.(2023)Troiano, Oberl{\"a}nder, and Klinger]{troiano-etal-2023-dimensional}
Enrica Troiano, Laura Oberl{\"a}nder, and Roman Klinger.
\newblock Dimensional modeling of emotions in text with appraisal theories: Corpus creation, annotation reliability, and prediction.
\newblock \emph{Computational Linguistics}, 49\penalty0 (1):\penalty0 1--72, March 2023.
\newblock \doi{10.1162/coli_a_00461}.
\newblock URL \url{https://aclanthology.org/2023.cl-1.1/}.

\bibitem[Yu et~al.(2026{\natexlab{a}})Yu, Yi, Karimi-Malekabadi, Abdurahman, Ye, Narayanan, Zhao, and Dehghani]{yu2026tracing}
Chenxiao Yu, Bowen Yi, Farzan Karimi-Malekabadi, Suhaib Abdurahman, Jinyi Ye, Shrikanth Narayanan, Yue Zhao, and Morteza Dehghani.
\newblock Tracing moral foundations in large language models.
\newblock \emph{arXiv preprint arXiv:2601.05437}, 2026{\natexlab{a}}.

\bibitem[Yu et~al.(2026{\natexlab{b}})Yu, Ma, Sun, Xu, and Zhang]{yu2026badcompany}
Wanying Yu, Boyang Ma, Zhibo~Eric Sun, Minghui Xu, and Yue Zhang.
\newblock Bad company corrupts good morals: Understanding and measuring narrative-induced moral reasoning degradation in llms.
\newblock \emph{arXiv preprint arXiv:2606.28981}, 2026{\natexlab{b}}.

\bibitem[Zellers et~al.(2019)Zellers, Holtzman, Bisk, Farhadi, and Choi]{hellaswag}
Rowan Zellers, Ari Holtzman, Yonatan Bisk, Ali Farhadi, and Yejin Choi.
\newblock {H}ella{S}wag: Can a machine really finish your sentence?
\newblock In \emph{Proceedings of the 57th Annual Meeting of the Association for Computational Linguistics}, pp.\  4791--4800, Florence, Italy, July 2019. Association for Computational Linguistics.
\newblock \doi{10.18653/v1/P19-1472}.
\newblock URL \url{https://aclanthology.org/P19-1472/}.

\bibitem[Zheng et~al.(2024)Zheng, Zhou, Meng, Zhou, and Huang]{Zheng0M0H24}
Chujie Zheng, Hao Zhou, Fandong Meng, Jie Zhou, and Minlie Huang.
\newblock Large language models are not robust multiple choice selectors.
\newblock In \emph{The Twelfth International Conference on Learning Representations, {ICLR} 2024, Vienna, Austria, May 7-11, 2024}. OpenReview.net, 2024.
\newblock URL \url{https://openreview.net/forum?id=shr9PXz7T0}.

\bibitem[Zou et~al.(2023)Zou, Phan, Chen, Campbell, Guo, Ren, Pan, Yin, Mazeika, Dombrowski, Goel, Li, Byun, Wang, Mallen, Basart, Koyejo, Song, Fredrikson, Kolter, and Hendrycks]{representation-engineering}
Andy Zou, Long Phan, Sarah~Li Chen, James Campbell, Phillip Guo, Richard Ren, Alexander Pan, Xuwang Yin, Mantas Mazeika, Ann{-}Kathrin Dombrowski, Shashwat Goel, Nathaniel Li, Michael~J. Byun, Zifan Wang, Alex Mallen, Steven Basart, Sanmi Koyejo, Dawn Song, Matt Fredrikson, J.~Zico Kolter, and Dan Hendrycks.
\newblock Representation engineering: {A} top-down approach to {AI} transparency.
\newblock \emph{CoRR}, abs/2310.01405, 2023.
\newblock \doi{10.48550/ARXIV.2310.01405}.
\newblock URL \url{https://doi.org/10.48550/arXiv.2310.01405}.

\end{thebibliography}
\end{document}